%% file: paper_revised.tex
\documentclass[sigconf]{acmart}

\AtBeginDocument{%
  \providecommand\BibTeX{{%
    \normalfont B\kern-0.5em{\scshape i\kern-0.25em b}\kern-0.8em\TeX}}}

\setcopyright{acmcopyright}
\copyrightyear{2023}
\acmYear{2023}
\acmDOI{10.1145/3588911}

\acmConference[SIGMOD '23]{the 2023 International Conference on Management of Data}{June 18--23, 2023}{Seattle, WA}

\acmPrice{15.00}
\acmISBN{978-1-4503-XXXX-X/18/06}

\keywords{Natural Language Question Answering, Knowledge Graphs, RDF,  Seq2Seq Models, Just-In-Time Entity and Relation linking }

\usepackage[english]{babel}
\usepackage{verbatim}
\usepackage{listings}
\usepackage{soul}
\usepackage{algorithm}
\usepackage{algpseudocode}
\usepackage{fancyvrb}
\usepackage{amsthm}

\usepackage{rotating}

\theoremstyle{definition}
\newtheorem{definition}{Definition}[section]

\colorlet{punct}{red!60!black}
\definecolor{background}{HTML}{EEEEEE}
\definecolor{delim}{RGB}{20,105,176}
\colorlet{numb}{magenta!60!black}

\def\ncp{\vspace*{0ex}}

\def\shorten{\looseness=-1}

\newcommand{\myNum}[1]{(\emph{#1})}
\newcommand{\sysName}{\textsc{KGQAn}}
\newcommand{\qald}{QALD-9}
\newcommand{\lcq}{LC-QuAD 1.0}
\newcommand{\dbpedia}{DBpedia}
\newcommand{\RDFTYPE}[3]{\ensuremath{{\langle}\texttt{#1},} \texttt{#2}, \ensuremath{\texttt{#3}{\rangle}}}
\newcommand{\nsstitle}[1]{\noindent\textup{\textbf{#1}}}

\newcommand{\glabel}[1]{\texttt{#1}}

\newcounter{example}[section]

\begin{document}



\author{Reham Omar$^\ast$, Ishika Dhall$^\ast$, Panos Kalnis$^\S$, Essam Mansour$^\ast$}
\affiliation{%
 \institution{ $^\ast$Concordia University, Canada \qquad\qquad\qquad $^\S$KAUST, Saudi Arabia}
\country{\qquad \texttt{\{fname\}.\{lname\}}@concordia.ca \qquad\qquad\quad panos.kalnis@kaust.edu.sa\qquad}
}

\title{A Universal Question-Answering Platform for Knowledge Graphs}

\input{sections/abstract}

\maketitle

\pagestyle{plain}

\input{sections/introduction}
\input{sections/comparative_analysis}
\input{sections/overview}
\input{sections/seq2seq}

\input{sections/jitLinking}

\input{sections/execution_manager}

\input{sections/results}

\input{sections/related_work}

\input{sections/conclusion}
\newpage
\bibliographystyle{ACM-Reference-Format}

\bibliography{references}

\end{document}

%% file: sections/abstract.tex
\begin{abstract}
\sloppy 

Knowledge from diverse application domains is organized as knowledge graphs (KGs) that are stored in RDF engines accessible in the web via SPARQL endpoints. Expressing a well-formed SPARQL query requires information about the graph structure and the exact URIs of its components, which is impractical for the average user. Question answering (QA) systems assist by translating natural language questions to SPARQL. 
Existing QA systems are typically based on application-specific human-curated rules, or require prior information, expensive pre-processing and model adaptation for each targeted KG. Therefore, they are hard to generalize to a broad set of applications and KGs. 
In this paper, we propose {\sysName}, a \emph{universal} QA system that does not need to be tailored to each target KG. Instead of curated rules, {\sysName} introduces a novel formalization of question understanding as a text generation problem to convert a question into an intermediate abstract representation via a neural sequence-to-sequence model.
We also develop a just-in-time linker that maps at query time the abstract representation to a SPARQL query for a specific KG, using only the publicly accessible APIs and the existing indices of the RDF store, without requiring any pre-processing. Our experiments with several real KGs demonstrate that {\sysName} is easily deployed and outperforms by a large margin the state-of-the-art in terms of quality of answers and processing time, especially for arbitrary KGs, unseen during the training. 

\end{abstract}

%% file: sections/introduction.tex
\section{Introduction}
 \label{sec:intro}
\sloppy
There is a steep rise in the availability of knowledge graphs (\emph{KG}s) across various application domains. Examples include KGs about scientific publications, such as the  Microsoft Academic Graph\footnote{\url{https://makg.org/}}, DBLP\footnote{https://dblp.org/} and OpenCitations\footnote{\url{https://opencitations.net/sparql}}; general-fact KGs, such as {\dbpedia}\footnote{\url{https://dbpedia.org/sparql}}, Wikidata\footnote{\url{https://query.wikidata.org/}} and YAGO\footnote{\url{https://yago-knowledge.org/sparql}}; knowledge related to bio-sciences (e.g., Bio2RDF\footnote{\url{https://bio2rdf.org/sparql}}); or politics (e.g., the UK Parliament\footnote{\url{https://api.parliament.uk/sparql}} KG). Due to the simplicity of the RDF model, the powerful SPARQL query language and the  extensions for inferencing and reasoning~\cite{BursztynGMR15,AMIE13}, many KGs are stored in RDF data management systems~\cite{lusail, AliSYHN22} that can be accessed remotely in the web as SPARQL endpoints.\shorten

\begin{figure}[t]
\begin{verbatim}
1 PREFIX dbv:<http://dbpedia.org/resource/> 
2 SELECT ?sea WHERE { 
3    ?sea <http://dbpedia.org/property/outflow>
          dbv:Danish_straits . 
4    ?sea <http://dbpedia.org/ontology/nearestCity> 
          dbv:Kaliningrad . }
\end{verbatim}
\caption{SPARQL query for question $q^E$. ``city on shore'' maps to \url{<http://dbpedia.org/ontology/nearestCity>}.}
\label{fig:nlq}
\end{figure}

Consider $q^E = $ ``Name the sea into which Danish Straits flows and has Kaliningrad as one of the city on the shore'', a question that appears in the popular {\lcq}~\cite{lcquad} benchmark. 
Intuitively, $q^E$  corresponds to a conjunctive SPARQL query consisting of two triples:    
\RDFTYPE{?sea}{flows}{DanishStraits} and
\RDFTYPE{?sea}{cityOnShore}{Kaliningrad}.
In practice, assuming the endpoint is {\dbpedia}, the query should be expressed as shown in Figure~\ref{fig:nlq}. Notice that the correspondence between elements of $q^E$ (e.g., ``city on shore'') and the actual SPARQL components (i.e., \url{<http://dbpedia.org/ontology/nearestCity>}) is not trivial.

It is impractical for the average user to know the schemata of the KGs, or the exact URIs of the entities and predicates to express the appropriate SPARQL queries. Hence, recent research \cite{gAnswer2014, gAnswer2018, EDGQA, NSQA} investigated the automatic translation of natural language questions to SPARQL; the problem is known as question answering (\emph{QA}). 
Existing QA systems utilize question understanding approaches based on semantic parsing and curated rules that are tailored to a specific KG; consequently, they are hard to generalize to broader application domains. 
Moreover, existing QA systems rely on prior knowledge for each target KG to link question phrases into KG vertices and predicates. For example, gAnswer~\cite{gAnswer2018, gAnswer2014} and EDGQA~\cite{EDGQA} perform expensive pre-processing to generate indices for each KG, whereas NSQA~\cite{NSQA} trains a specialized deep neural network on the target KG. This has several drawbacks: \myNum{i} the computational cost of pre-processing is substantial, especially for large KGs; \myNum{ii} QA is only available at those SPARQL endpoints whose owners choose to install and  continuously maintain the QA system; and \myNum{iii} it is hard to generalize to arbitrary KGs.

\begin{figure}[t]
\vspace*{-1ex}
  \centering
    \includegraphics[width=\columnwidth]{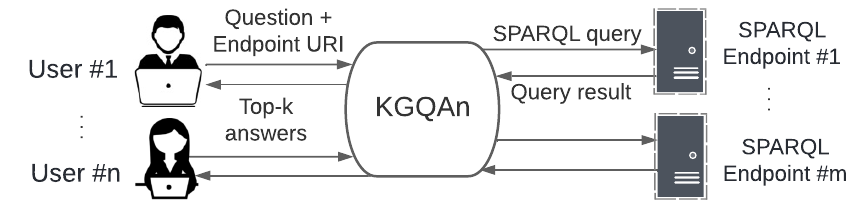}

\caption{Users interact on-demand with {\sysName} to submit a natural language question against an arbitrary KG. Question understanding is independent of the  KG. The construction of the actual SPARQL query is performed in a JIT manner, through the publicly accessible API of the target KG.}  
\label{fig:ondemandQAS}
\vspace*{-3ex}
\end{figure}

In contrast to existing systems, we are inspired by web search engines that resolve user questions independently of any particular web site. We propose a novel approach, called {\sysName}\footnote{{\sysName}: \underline{K}nowledge  \underline{G}raph \underline{Q}uestion \underline{AN}swering platform.},  that forgoes the need of tailoring the QA system to each individual KG. {\sysName} resides between the users and any available KG, as shown in Figure~\ref{fig:ondemandQAS}. 
{\sysName} introduces novel approaches for  \emph{question understanding}, \emph{linking}, and \emph{filtering} of the final answers.
We formalize question understanding as a text generation task that extracts abstract triple patterns from the natural language question. Prior to deployment, we train {\sysName} on a diverse set of questions using a sequence-to-sequence (Seq2Seq)~\cite{seq2seq} deep neural network. The resulting model takes a question as input and extracts a set of triple patterns that represent the relations between entities, which are either explicitly mentioned, or are unknowns. With those triple patterns we construct the so-called phrase graph pattern (\emph{PGP}), which is a formal abstract representation of the system's understanding of the question, independently of any KG.

We also propose a just-in-time (\emph{JIT}) linker that negotiates with the target KG the mapping between abstract elements of the PGP and the appropriate URIs to instantiate a well-formed SPARQL query. {\sysName} submits requests via the publicly accessible SPARQL endpoint API and utilizes common built-in indices \cite{quoc2019performance} that exist in all popular RDF stores (e.g., Virtuoso, Stardog, and Apache Jena). Linking is performed transparently via a semantic affinity model trained on general English text, without the need of any particular configuration or action by the owner of the KG. 
The SPARQL query is submitted to the KG and the answers are collected and filtered by {\sysName} before being returned to the user. Filtering is based on constraints such as the expected domain, or data type of the answer; the constraints are identified during the construction of the PGP. Unlike existing QA systems, our filtering method does not need any prior knowledge about the KG.\shorten 

We evaluate {\sysName} on a variety of real KGs. In addition to {\dbpedia}, which is used by the two popular QA benchmarks, {\lcq}~\cite{lcquad} and {\qald}~\cite{qald9}, we also experiment with  YAGO, the Microsoft Academic Graph and DBLP.  
In terms of answer quality (i.e., Macro F1 score) and question processing time, our experiments confirm that {\sysName} outperforms existing QA systems by a large margin when queries are executed on arbitrary KGs, unseen during training. Interestingly, despite the fact that {\sysName} is general and does not target any particular KG, it manages to perform almost identically, or in some cases even outperform, the current state-of-the-art (\emph{SOTA}) on the {\lcq} and {\qald} benchmarks; note that the SOTA is optimized specifically for those benchmarks. 


In summary, our contributions are:
\begin{itemize}

\item {\sysName}, the first universal QA system that can  process on-demand questions against arbitrary, previously unseen KGs.

\item  A novel formalization of question understanding as a text generation problem. We train Seq2Seq neural networks that learn how to extract a formal abstract representation from the natural language question, independently of the application domain and without any curated rules.\shorten

\item A novel just-in-time linking and filtering approach that utilizes a generic semantic affinity model and the public API of the SPARQL endpoint.  Our approach does not need prior knowledge of the KG and aims at maximizing the recall of the SPARQL query, before improving  precision via filtering.

\item A comprehensive evaluation using four real KGs from different domains. {\sysName} performs similarly to the SOTA for the known benchmarks, but outperforms the SOTA by a large margin for previously unseen KGs.


\end{itemize}

The paper is organized as follows: 
Section~\ref{sec:comp} clarifies the necessary background and limitations of existing systems. 
Sections~\ref{sec:overview}, \ref{sec:triples}, \ref{sec:linking} and \ref{sec:exec} present the architecture of {\sysName}, as well as our query understanding, linking and filtering approaches. Section~\ref{sec:results} contains the experimental evaluation.  
Section~\ref{sec:relatedwork} discusses the related work and Section~\ref{sec:conc} concludes the paper.

%% file: sections/comparative_analysis.tex
\section{Background and Limitations}
\label{sec:comp}

QA systems commonly split the question answering process into three steps: 
\myNum{i} \emph{question understanding}: extracts the entities and relations from the question and generates an  abstract representation;  \myNum{ii} \emph{linking}: maps the abstract representation to the corresponding vertices and predicates in the targeted KG to construct a well-formed SPARQL query; and \myNum{iii} \emph{filtering}: selects the most relevant answers, based on some constraints.   
In this section we discuss three existing QA systems, gAnswer~\cite{gAnswer2014, gAnswer2018}, EDGQA~\cite{EDGQA} and  NSQA~\cite{NSQA}, summarized in Table~\ref{tab:background}. These systems highlight the most successful approaches, as they represent the SOTA in terms of accuracy (i.e., F1 score) on the {\lcq} and {\qald} benchmarks.

\input{tables/comparison}

\subsection{Question Understanding}
\label{sec:BQU}

\begin{figure}
  \centering
    \includegraphics[width=\columnwidth]{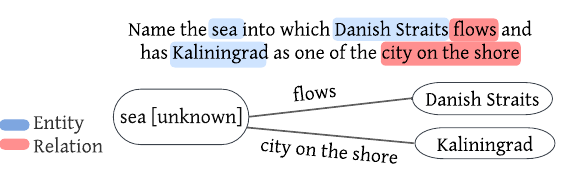}

\caption{The understanding of question $q^E$ identifies \glabel{Danish Straits} and \glabel{Kaliningrad} as  mentioned entities; \glabel{sea} as an unknown, \glabel{flows} and \glabel{city on the shore} as relations, resulting in a graph with two triple patterns. 
}
\label{fig:understanding}
\end{figure}

The question understanding step identifies mentioned entities, unknowns and relations in the natural language question. It also generates an abstract formal representation (typically, a graph) of their inter-dependencies. Figure~\ref{fig:understanding} shows the expected result for our running example $q^E$.  

\textbf{gAnswer} 
uses the Stanford dependency parser~\cite{stanforddependency} to map a question into a syntactic dependency tree. The tree is traversed to generate an abstract graph representation of the query, called semantic query graph. This is achieved by a set of static heuristic linguistic rules that are curated on the set of the {\qald} training questions; for instance,  question words starting with ``wh*'' (e.g., ``what'') are assumed to represent unknowns. gAnswer also depends on a predefined set of synonyms~\cite{Syn}.
For our running example, gAnswer 
generates a lot of unnecessary and erroneous triples, such as \RDFTYPE{Kaliningrad}{city of}{city}.


\textbf{EDGQA} utilizes the Stanford core NLP parser~\cite{CoreNLP}
to construct the constituency parsing tree of the question. Then, it generates a rooted acyclic graph, called entity description graph, by iteratively decomposing the parsing tree. The process is based on rules that are tailored to the {\lcq} and {\qald} benchmarks. 
For our running example, EDGQA erroneously detected~\cite{EDGQALog}
\glabel{Danish Straits} as a relation, \glabel{flow} as an entity, and generated this triple pattern: \RDFTYPE{unknown}{Danish Straits}{flow}.



\textbf{NSQA} parses the question into a directed rooted acyclic graph, called abstract meaning representation. Parsing is based on a deep neural symbolic framework. 
The neural network is trained  on verb-oriented relations gathered from Propbank. The training set also consists of a set of curated trees for different questions. 
The accuracy of the model depends on the diversity of these trees. It is unclear\footnote{NSQA is a proprietary IBM project, not available to us to reproduce the results, or test our running example $q^E$; our discussion is based on information from \cite{NSQA}.} whether the model can detect relations based only on noun phrases, such as ``city on shore''. 
The resulting graph has different granularity than SPARQL; thus, the model has to be adjusted \cite{NSQA} to work with new domains.\shorten

\subsection{Entity and Relation Linking}
\label{sec:BLinking}

In QA systems, linking is a challenging task~\cite{LinLXLC20} that attempts to map entities and relations extracted from a question to the corresponding vertices and predicates in a KG. For our running example $q^E$ on {\dbpedia}, the expected output is:
\input{tables/expectedLinking}

Existing systems formalize linking as:
\myNum{i} a lookup in an inverted index that maps synonyms to relevant vertices and predicates of the target KG;
or \myNum{ii} a keyword search on a database of tagged documents constructed from the vertices and predicates of the target KG;   
or \myNum{iii} a deep learning-based transformation, defined on the vector space of the target KG. 
These methods are KG-specific and require expensive pre-processing for each target KG.


\textbf{gAnswer} links entities via the crossWikis dictionary~\cite{gAnswerDict}, constructed from a web crawl. Each entity in the dictionary is mapped to a corresponding vertex in the target KG with a confidence score.
For relation linking, gAnswer constructs a 
dictionary that maps relation mentions to their corresponding predicates in the target KG. For example, relation ``starred by'' is mapped to predicate \glabel{starring} in a particular KG. 
The mapping is generated by general NLP~\cite{PATTY} methods, or by using n-grams~\cite{FaderSE11}.\shorten 

\textbf{EDGQA} utilizes three indexing systems, namely Falcon~\cite{falcon}, EARL~\cite{earl} and Dexter \cite{dexter}, to index vertices and predicates of the target KG. EDGQA ensembles the results from the three systems to link entities to KG vertices; then uses those to limit the search space for relation linking~\cite{eearl}; and finally utilizes a BERT-based~\cite{bert} model to rank the list of predicates. 
The aforementioned indexing systems process a target KG by extracting vertices and predicates with their associated descriptions. 
For example, Falcon uses standard predicates \glabel{\#label} and \glabel{\#altLabel}
to collect  descriptions and synonyms.
On those, Falcon performs part-of-speech tagging and  n-gram extraction to generate tagged documents, in order to enable keyword search during linking. 
If a KG does not contain \glabel{\#label} predicates, an appropriate indexing predicate must be selected manually for each vertex type, e.g., \glabel{authorName} can be chosen to index vertices of type \glabel{author}.


\textbf{NSQA} is slightly different, since it identifies entities and relations during linking. First, it employs Blink \cite{blink} to generate an embedding of the target KG vertices in a vector space. 
Then, it receives the abstract graph from the previous phase and detects entities via a BERT-based neural model, trained on 3,651 questions from {\lcq} with manually annotated mentions. The detected entities are embedded in the same space as the KG and are linked to their nearest-neighbor vertices. 
For relation linking, NSQA uses SemRel~\cite{nsqa_linking}, a transformer-based neural model, which is trained to predict the relation. The model returns a list of candidate relations from the set of relations in the KG and ranks them. In NSQA, building the vector space, that is, generating the embedding for the target KG vertices and predicates, dominates the cost of the pre-processing phase. 


\begin{figure*}[t]
\vspace*{-1ex}
  \centering
  \includegraphics[width=\textwidth]{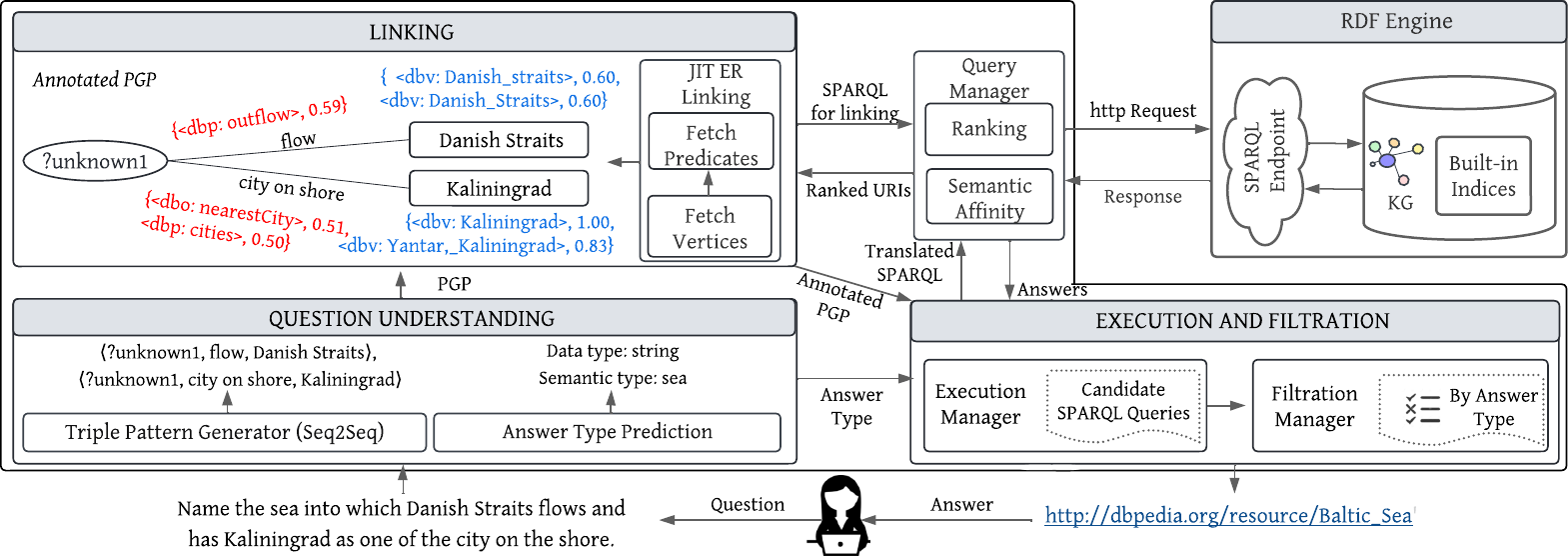}
  \caption{
{\sysName}'s architecture includes three phases: a) Question Understanding: a PGP is constructed using our Triple Pattern Generator (seq2seq model), b) Linking PGP to KG: JIT ER linking module annotates the PGP with relevant vertices and predicates from a KG, and c) Execution and Filtration: creating and executing SPARQL queries, and post-filtering their results.
}
\label{fig:architecture}
\vspace*{-3ex}
\end{figure*}

\subsection{Filtering}
Some QA systems, such as EDGQA, include a step for filtering the potential answers by the RDF engine. For the filtering step, existing systems build prior knowledge during a pre-processing phase of all node types/classes in a KG. The filtering is done by linking unknown (e.g., \glabel{?sea}) to a vertex type in the target KG, such as \RDFTYPE{?sea}{a}{dbo:Sea}, if such a type exists. The SPARQL query amended with the type constraint is sent for execution; therefore, the received answers are pre-filtered. 


%% file: tables/comparison.tex
\begin{table*}[t]
\vspace*{-1ex}
  \caption{A Comparative Analysis of SPARQL-based QA systems achieved highest F1 scores in QALD-9 and/or LC-QuAD 1.0}
\vspace*{-2ex}
   \label{tab:background}
    \begin{tabular}{lccc}
        \hline
        & \textbf{Question Understanding}&\textbf{Linking}& \textbf{Filtering} \\
        \hline
        gAnswer~\cite{gAnswer2014, gAnswer2018} & Dependency parsing &  Index look up  & n/a \\
        EDGQA~\cite{EDGQA} & Constituency parsing & Dexter, EARL, Falcon & By index type\\
        NSQA~\cite{NSQA} & AMR parsing, BERT-based & BLINK, SEM-Rel & n/a  \\
        \textbf{{\sysName} \textbf{(ours)}} & Seq2Seq text generation & RDF engine + Semantic similarity & By predicted answer type\\
        \hline
    \end{tabular}
   \vspace*{-1ex}
\end{table*}

%% file: tables/expectedLinking.tex
\begin{table}[h] 
\ncp\ncp
  \ncp\ncp\ncp
  \label{tab:expectedLinking}
    \begin{tabular}{ll}
        \hline
        \textbf{Phrase}&{\textbf{URI}} \\
        \hline
        Danish Straits & <\url{http://dbpedia.org/resource/Danish\_straits}>\\
        Kaliningrad & <\url{http://dbpedia.org/resource/Kaliningrad}>\\
        flows&<\url{http://dbpedia.org/property/outflow}>\\
        city on the shore & <\url{http://dbpedia.org/ontology/nearestCity}>\\ \cline{1-2}\hline
    \end{tabular}
\ncp\ncp\ncp\ncp
\end{table}

%% file: sections/overview.tex
\section{{\sysName} Architecture}
\label{sec:overview}

Similar to existing approaches, {\sysName} implements three distinct phases: question understanding, linking, and execution with filtration, as shown in Figure~\ref{fig:architecture}. However, in contrast to existing approaches that require prior knowledge and expensive pre-processing for each target KG, {\sysName} is a universal solution that works with arbitrary KGs from a variety of application domains.


\textbf{Question understanding.} {\sysName} formalizes question understanding as a text generation task. We employ a pre-trained Seq2Seq language model, such as BART~\cite{BART} or GPT-3~\cite{gpt}. The original model understands general English language text; we amend its training with a set of typical questions and their corresponding translation to abstract triple patterns. Training is performed independently of any KG. Consequently, the model accepts general English questions and generates a sequence of abstract triple patterns. Figure~\ref{fig:architecture} shows that, for our running example $q^E$, the seq2seq model generates two triples:   
\RDFTYPE{?unknown1}{flow}{Danish Straits} and
\RDFTYPE{?unknown1}{city on shore}{Kaliningrad}.
The set of all triples correspond to an abstract graph representation, called phrase graph pattern (\emph{PGP}).\shorten


{\sysName} also predicts the expected data and semantic type of the answer. 
We train a three-layer neural network; its input is the question in English and the output is the expected data type of the answer, that is, date, numerical, boolean, and string. In the latter case, we also predict the semantic type, based on the context. For our example question, the predicted data type is string, whereas the predicted semantic type is ``sea''.


\textbf{Linking.} Intuitively, linking corresponds to semantic match between elements of the abstract PGP graph and the actual data in the target KG. {\sysName} does not have any prior information about the target KG; therefore, the match must be performed in a just-in-time (JIT) manner, during question answering. This allows {\sysName} to support arbitrary KGs.   
Since most RDF engines do not support semantic search, we split this process between {\sysName} and the RDF engine. {\sysName} submits, through the standard SPARQL API, a set of text containment queries for the phrase associated with each entity in the PGP; these are answered by the built-in indices \cite{quoc2019performance} of the RDF engine. The answers, which are URIs of potential semantic matches to the PGP phrases, are assigned  by {\sysName} a semantic affinity score, based on an existing generic word embedding model. The top-$k$ matches are then used to query again the target KG (by a standard SPARQL query), this time in order to acquire potential semantic matches for the predicates of the PGP. The result of this process is an \emph{annotated} PGP. For our running example in Figure~\ref{fig:architecture}, entity \glabel{Kaliningrad} has two matches, \glabel{dbv:Kaliningrad} with score $1.00$ and \glabel{dbv:Yantar,\_Kaliningrad} with score $0.83$; whereas relation \glabel{flow} has one match, \glabel{dbp:outflow} with score $0.59$.


\textbf{Execution and filtration.} 
{\sysName} traverses the annotated PGP to generate all combinations of well-formed candidate SPARQL queries that can be semantically equivalent to the user question, for the target KG. The top-$k$ most promising queries, based on a semantic affinity score, are sent to the RDF engine. The answers are collected by  {\sysName} and are filtered based on the aforementioned expected data and semantic type. Note that, while the execution of multiple candidate queries increases recall, the filtering step crutially improves precision. 
In contrast to existing systems that embed a KG-specific filter within the SPARQL query, our filtering method is applied at a post-processing phase, independently of the KG; therefore, our method is applicable to arbitrary KGs. For our runing example, the user receives \glabel{http://dbpedia.org/resource/Baltic\_Sea} as final answer.\shorten


%% file: sections/seq2seq.tex
\section{{\sysName} Question Understanding}
\label{sec:triples}

This section introduces our novel formalization of the question understanding problem as a text generation task.
Given a question $q$ in plain English, we generate a sequence of triple patterns 
\RDFTYPE{entity$^a_i$}{relation{$_{i}$}}{entity$^b_i$} that represent our formal understanding of $q$; this is achieved by a Seq2Seq deep neural network. All components of the triples are either phrases from $q$, or unknowns (i.e., variables); therefore question understanding is independent of any specific domain or KG. Unknowns, in particular, are first-class citizens; we detect their semantic relations with mentioned entities, and predict their expected data and semantic type.   
Note that, our work is related to 
relation triple extraction \cite{triples2020, triples2022} that refers to the construction of a KG from long text; however, the existence of unknowns renders our task more challenging.


\subsection{Phrase Triple Patterns Extraction}
\label{sec:triples_task}
Question $q$ in plain English consists of a sequence of words ($q_1, q_2, q_3, ... $), where $q_i$ is the $i$th word. Our task is to generate a sequence of triples that capture the semantics of $q$. Formally:


\begin{definition}[Phrase Triple Pattern Extraction]
\label{def:pgp} 
Let $q = (q_1, q_2, q_3, ... )$ be a question.
Generate a sequence of triples $TP(q) = (tp_1, tp_2, ...)$. Each $tp_i \in TP(q)$ has the form $tp_i =$
\RDFTYPE{e$^a_i$}{r{$_i$}}{e$^b_i$}, where
$e^a_i$ and $e^b_i$ are either entity phrases in $q$, or unknowns; and $r_i$ is a relation phrase in $q$.
\end{definition}

%

\subsubsection{Triple extraction by a Seq2Seq transformer. }

We model our text generation task using a Seq2Seq pre-trained language model (\emph{PLM}). 
The input query $q$ is in plain text.
For compatibility with the seq2seq model, the output $TP(q)$ should also be represented as text \cite{seq2seq}. 
Let $y=F_{txt}(TP(q))$ be the text representation. $F_{txt}$ transforms each $tp_i \in TP(q)$ into plain text by annotating $e^a_i, e^b_i, r_i$ as ``EntityA'', ``EntityB'' and ``Relation'', respectively. Our example query $q^E$ results in two triples:  
$TP(q^E)=$
(\RDFTYPE{?unknown1}{flow}{Danish Straits},
\RDFTYPE{?unknown1}{city on shore}{Kaliningrad}), that correspond to the following triple patterns:

\begin{table}[h]
\begin{tabular}{l}
\hline
\texttt{[Relation(label="flow"),} \\
\texttt{\hspace{8pt}EntityA(label="Name", category=variable, varID=1),}\\
\texttt{\hspace{8pt}EntityB(label="Danish Straits", category=entity)],} \\
\texttt{[Relation(label="city on shore"),}\\
\texttt{\hspace{8pt}EntityA(label="Name", category=variable, varID=1),} \\
\texttt{\hspace{8pt}EntityB(label="Kaliningrad", category=entity)]}\\ 
\hline
\end{tabular}
\end{table}

\begin{figure}
  \centering
  \includegraphics[width=\columnwidth]{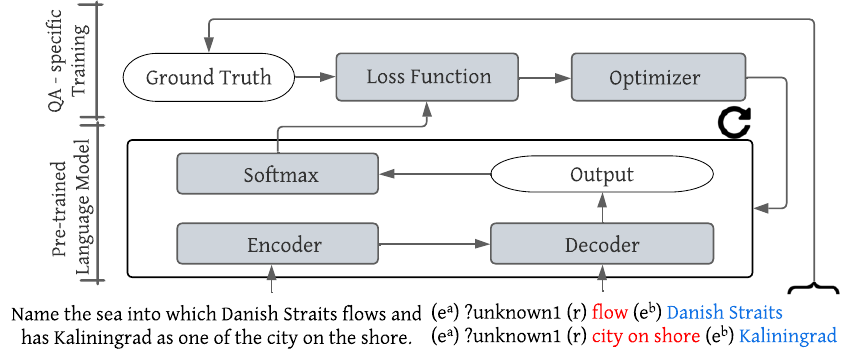}
  \caption{Training our Seq2Seq model with a question and a sequence of triple patterns as inputs to the encoder and decoder, respectively. Training is based on 1,752 manually annotated questions.  
}
\label{fig:Seq2Seq}
\end{figure}

We consider two categories of PLMs: \myNum{i} encoder-decoder, such as BART~\cite{BART};
and \myNum{ii} decoder-only, such as GPT-3~\cite{gpt}. These PLMs support dynamic-length sequence generation~\cite{PLMAd}.
Figure~\ref{fig:Seq2Seq} illustrates our training process using an encoder-decoder PLM. The encoder receives $q$ as input and applies a self-attention mechanism to generate a vector representation (embedding) of $q$. The encoder-embeddings $F^e(q)$  encapsulate various features of $q$ that emphasize its most important tokens. The decoder learns to generate triple patterns as a sequence of tokens, where predicting token $i$ is based on $F^e(q)$ and all previous tokens. The decoder uses two attention mechanisms: \myNum{i} self-attention on its input $y$, and \myNum{ii} cross-attention between $F^e(q)$, and $y$~\cite{transformer}. If a decoder-only PLM is used, $q$ is directly provided as input to the decoder, which is trained to generate $y$ using only a self-attention mechanism on  $q$~\cite{gpt}. 

\subsubsection{Training dataset for triple pattern extraction.} 

The Seq2Seq model, either BART or GPT-3, is pre-trained with general English text. To be used for question understanding, we must perform additional training for triple extraction, as shown in Figure~\ref{fig:Seq2Seq}. To achieve this, we prepare a manually annotated training dataset using $1,752$ questions collected from the training datasets of {\lcq}~\cite{lcquad} and {\qald}~\cite{qald9}. Each question $q$ is annotated with an appropriate set of triples $TP(q)$. For each $q$, the main steps of the annotation process are:

\begin{itemize}
    \item Identify the number of phrase triple patterns $|TP(q)|$.
    \item Detect the triple patterns that share the same unknown.
    \item Assign a unique identifier for each unknown. 
    \item For each \RDFTYPE{e$^a_i$}{r{$_i$}}{e$^b_i$}, $\in TP(q)$, extract from $q$ the entity and relation phrases that correspond to $e^a_i$, $e^b_i$ and $r_i$, respectively.  
\end{itemize}

We design our training dataset
to include questions with different numbers of unknowns and triple patterns. In our annotation, we assume one main unknown\footnote{Both {\qald} and {\lcq} benchmarks do not include questions with two intentions. In future work, we plan to extend {\sysName} to support questions with two intentions, e.g., when and where did Covid-19 start?} (i.e., intention) for which we need to provide an answer. Additional unknowns may exist, but are treated as intermediate variables. Distinguishing between main and intermediate unknowns helps {\sysName} at the execution and filtration phase. For example, intermediate variables can be used in the WHERE clause, but not in the SELECT clause of the generated SPARQL query.\shorten

We consider various cases for the  syntactic and semantic expression of entities and relations. For example, an entity may be a named entity, such as ``Danish Straits''; or an entity mention, such as ``capital region''. A relation may be a verb, such as ``flows''; a verb with adverb, such as ``work out''; or a noun phrase, such as ``city on shore''. Our dataset allows the Seq2Seq model to learn how to distinguish between entity and relation phrases, without any additional semantic parsing, or part-of-speech tagging.   

Our dataset can be amended in the future with additional representative cases. However, it is already general enough to allow the Seq2Seq model to extract triples from a variety of domains. For example, although our training involves general facts from {\dbpedia}, without any questions about publications, it can understand questions related to DBLP, as shown in our experimental evaluation. Also note that the annotated dataset does not contain the actual SPARQL query, since this would depend on the target KG, which is not known during training.

\subsection{Phrase Graph Pattern (PGP)}
Let $q$ be a question and $TP(q)$ be the corresponding set of triple patterns. We define the phrase graph pattern for $q$ as:

\begin{definition}[Phrase Graph Pattern (PGP)]
\label{def:pgp} 
Let $\mathcal{E}, \mathcal{R}$ be a set of nodes and edges, respectively. The phrase graph pattern of $q$ is an undirected graph $PGP(q)=(\mathcal{E}, \mathcal{R})$, such that:  
for every triple \RDFTYPE{entity$^a_i$}{relation{$_i$}}{entity$^b_i$} $\in TP(q)$
there is a node $e^a_i \in \mathcal{E}$ with label $\texttt{entity}^a_i$;
a node $e^b_i \in \mathcal{E}$ with label $\texttt{entity}^b_i$;
and an edge $(e^a_i,e^b_i) \in \mathcal{R}$ with label $\texttt{relation}_i$.
\end{definition}



Intuitively, $PGP(q)$ connects the set of triple patterns in $TP(q)$ into a graph that represents the formal understanding of question $q$, as shown in Figure~\ref{fig:architecture}. Note that, although RDF data form directed graphs, $PGP(q)$ is undirected. The reason is that the PGP is not aware of the target KG. Therefore, at this point, we do not know yet the predicates that appear in the target KG, nor their direction.
Depending on the question, the constructed PGPs may form different shapes, such as star, or path query graphs. 
Following the different categories of questions in~\cite{lcquad2}, the current implementation of {\sysName} can support a wide range of questions, including: single fact, single fact with type, multi-fact, and Boolean questions.

\subsection{Data and Semantic Type Prediction}
As explained above, the resulting PGP contains one main unknown. {\sysName} predicts the expected data type and the semantic type of the unknown; this information will be used after the query execution, in order to improve the precision of the answers. {\sysName} uses only the user question $q$ to predict the data and semantic types, irrespectively of the target KG.

The expected data type can be date, numerical, boolean, or string. We define data type prediction as a classification task. We train a deep neural network  using the training dataset in the {\qald} benchmark, since its questions are already annotated with the data type. 
If the expected data type is string, then we also predict the semantic type (e.g., person, book, etc). We use the following heuristic: the  
first noun in the question is the semantic type. 
We utilize the  AllenNLP constituency parser~\cite{allennlp} to acquire the part-of-speech tags in the question, and extract the first phrase whose tag is noun. This heuristic has impact only on the accuracy of {\sysName}'s post-filtering step; question understanding and linking are not affected.
For our running example $q^E$, {\sysName} predicts the data and semantic types to be string and ``sea'', respectively.

%% file: sections/jitLinking.tex
\section{{\sysName} Linking}
\label{sec:linking}

For a question $q$, linking receives $PGP(q)$ from the question understanding phase, and generates an annotated version $AGP(q)$ of the graph. Vertices and edges in $AGP(q)$ are annotated with the URIs from the target KG that are semantically closest to the meaning of $q$. Since the target KG is not known in advance, {\sysName} implements linking at query time, in a just-in-time manner using a set of light-weight SPARQL requests. This allows {\sysName} to work with arbitrary KGs, without any pre-processing.

\subsection{Entity Linking}

Let $l_n$ be the label of a node $n$ in $PGP(q)$. Let $d_v$ be the description (i.e., a literal of type string) of a vertex $v$ in the target knowledge graph $KG$. 
Let $S(l_n, d_v)$ be the semantic affinity score between $l_n$ and $d_v$; refer to Section~\ref{sec:semanticAffinity} for the definition of $S(\circ, \circ)$. Intuitively, for each node $n$ we need to go though all vertices $v$ in $KG$; compute their semantic affinity to $n$; and select the top-$k$ matches. These are the \emph{relevant} vertices that are added as  annotation on $n$ in the annotated graph $AGP(q)$. 

There are two practical issues: First, how to extract the description $d_v$ of node $v$ in $KG$. Some KGs have human-readable URIs and clearly marked descriptions via the \glabel{rdf:label} predicate. For example, the entry for Princess Diana in {\dbpedia} contains triple \RDFTYPE{http://dbpedia.org/resource/Diana,\_Princess\_of\_Wales}{rdf:label}{``Diana, Princess of Wales''}. We can check with a SPARQL ASK query if $KG$ contains \glabel{rdf:label} predicates, in which case we can directly retrieve the descriptions. Other KGs, however, are more cryptic. For example, the entry for Jim Gray in the Microsoft academic graph, is \glabel{mag:2279569217}. Thankfully, there is a triple \RDFTYPE{https://makg.org/entity/2279569217}{foaf:name}{``Jim Gray''} from which we can retrieve the description. However, \glabel{foaf:name} is an arbitrary predicate that may differ for other entities. Therefore, in general we retrieve triples that connect $KG$ vertices to literals of type string via any predicate. 

The second issue is that typical RDF engines do not support semantic search. Consequently, function $S(\circ, \circ)$ must be computed remotely, at the {\sysName} site. Obviously, we do not want to transfer a large amount of (possibly irrelevant) data from $KG$ to {\sysName}. We use the following heuristic: label $l_n$ consists of a sequence of words (e.g., ``Danish Straits''). We request from $KG$ only those vertices $v$ that are connected with a literal $d_v$ via any predicate $p$, such that $d_v$ contains any combination of words from $l_n$. The heuristic translates to the following SPARQL query, where $Q(l_n)$ represents a disjunctive boolean expression for all words in $l_n$:

\begin{figure}[h]
\begin{verbatim}
QUERY: potentialRelevantVertices(l_n, maxVR)
   1 SELECT DISTINCT ?v ?d_v 
   2 WHERE { 
   3    ?v ?p ?d_v .   
   4    ?d_v <bif:contains> Q(l_n) . }
   5 LIMIT maxVR 
\end{verbatim}
\end{figure}

\noindent All modern RDF engines, such as Virtuoso, Stardog, and Apache Jena, construct by default full-text indices to enable text search~\cite{quoc2019performance}; 
therefore, \glabel{potentialRelevantVertices} can be executed efficiently. Since the query may still return numerous matches, we heuristically limit the result size to $maxVR$; in our experiments, we set $maxVR=400$. 
The query assumes Virtuoso as the RDF engine. Other engines may expose a slightly different API; for example, for Stardog we replace \glabel{<bif:contains>} with \glabel{<stardog:textMatch>}.
We formally define the set of relevant vertices as:

\begin{definition}[Relevant vertices $\mathcal{R_V}$]
\label{def:rVertex}
Given a question $q$ and a target knowledge graph $KG= (V, P)$ considered as a set of triples.
Let $n$ be a node in $PGP(q)$ with label $l_n$.
Let $T_d = \left\{\RDFTYPE{v}{p}{$d_v$}: v\in V, p\in P \right\}$ be a subset of triples from $KG$, such that description $d_v$ is a literal of type string, and  $l_n$ is fully or partially contained in $d_v$. 
Let $T_v = \left\{\langle v,S(l_n,d_v)\rangle: \RDFTYPE{v}{p}{$d_v$} \in T_d \right\}$, where $S(\circ, \circ)$ represents the semantic affinity score.
The set of relevant vertices $\mathcal{R_V}(n, KG)$  for node $n$ in $KG$, is the subset of $T_v$
that contains all pairs with the top-$k$ affinity scores. 
\end{definition}


Observe that the \glabel{potentialRelevantVertices} query, which is executed at the target KG, is a heuristic that retrieves fast a relatively large and potentially inaccurate set of vertices. For PGP node \glabel{Kaliningrad} in our running example $q_E$, it returns both \glabel{dbv:Kaliningrad} and \glabel{dbv:Yantar,\_Kaliningrad} (see Figure~\ref{fig:architecture}). The semantic affinity score, which is computed at the {\sysName} site, identifies \glabel{dbv:Kaliningrad} as the top match.

\input{algorithms/entity_linking_algorithm}

Algorithm~\ref{alg:entityLinking} summarizes the entity linking process for a single node $n$; the algorithm must be called for every node  $n \in PGP(q)$. In line 1 we check if $n$ is an unknown (i.e., variable), in which case there will be no relevant vertices at this phase (line 2). Line 4 executes the \glabel{potentialRelevantVertices} SPARQL query at the target RDF engine. Lines 5-8 compute for each returned vertex its semantic affinity to $n$, and store the resulting $\langle vertex, score \rangle$ pairs in $T_v$, as explained in Definition~\ref{def:rVertex}. Finally, line 9 selects from $T_v$ the pairs with the top-$k$ score to construct the set of relevant vertices $\mathcal{R_V}(n, KG)$.


\subsubsection{Complexity} 
Algorithm~\ref{alg:entityLinking} is executed $|\mathcal{E}|$ times, where $\mathcal{E}$ is the set of nodes in $PGP(q)$; see Definition~\ref{def:pgp}. 
The cost is dominated by the SPAQRL query at line 4.  
Let $c_{rdf}$ be the cost of performing keyword search in an RDF engine. $c_{rdf}$ is determined by the indices and  methods implemented in the particular engine. For example, Apache Jena implements full text search by either Lucene~\cite{jenaTS}, or Elastic search. 
The for-loop in line 6 is executed a constant number of times, because it is constrained by parameter $maxVR$ in the SPARQL query. The cost of the semantic affinity calculation in line~7 depends on $|l_n|\cdot |d_v|$. However, in practice both of these sets contain only a few words, so the cost can be considered constant. The top-$k$ computation in line 9 is also constrained by $maxVR$, so it is constant.      
The resulting complexity of Algorithm~\ref{alg:entityLinking} is $O(c_{rdf}|\mathcal{E}|)$.

\subsection{Relation Linking}

Let $p$ be a predicate in the target knowledge graph $KG$.
Since $KG$ is directed, $p$ must be connected to some vertex $v \in KG$ either as an outgoing, or incoming edge. If we consider $KG$ in its triple representation, there exists triple \RDFTYPE{$v$}{$p$}{?obj}, or \RDFTYPE{?sub}{$p$}{$v$}, respectively. 
Let $d_p$ be the description (i.e., a human-readable string) of $p$. In some KGs, the URIs of the predicates contain human-readable text, such as \glabel{dbo:spouse} in {\dbpedia}. In this case, we assume the URI to be the description of the predicate, i.e., $d_p = p$. We issue two different SPARQL queries to get outgoing and incoming predicates, with respect to vertex $v$. The SPARQL query to retrieve $d_p$ for an outgoing predicate $p$ is:\shorten

\begin{figure}[h]
\begin{verbatim}
QUERY: outgoingPredicate(v)
  1 SELECT DISTINCT ?p  
  2 WHERE { 
  3    v ?p ?obj .}
\end{verbatim}
\end{figure}



Query \glabel{incomingPredicate(v)} is exactly the same, except from line 3, which becomes \texttt{?sub ?p v}. Both outgoing and incoming queries can be executed efficiently by most RDF engines that index all triples in six ways~\cite{Hexastore, TripleBit} for traditional lookup.
In some KGs, however, the predicate URIs might be arbitrary internal identifier\footnote{Wikidata uses this naming system and provides a method to query the description, e.g., \url{https://www.wikidata.org/wiki/Wikidata:SPARQL\_query_service/queries\#Adding\_labels\_for\_properties}}. For example, a predicate with URI $p=$ \glabel{wdg:P227}, where the description of \glabel{wdg:P227} could also be retrieved by a SPARQL query from the KG \glabel{wdg}. After receiving the results of our queries, we check: If $?p$ is an arbitrary internal identifier, then we issue another query to get the predicate description.


Let $r$ be a relation in $PGP(q)$, connecting two nodes $n^a, n^b \in PGP(q)$. The nodes are already linked to relevant vertices $\mathcal{R_V}(n^a, KG)$ and $\mathcal{R_V}(n^b, KG)$, respectively. Our goal is to link $r$. The intuition of our approach is that any successfully linked pair $\langle node, relation\rangle$ in the PGP must occur at least once in the target KG. Therefore, instead of blindly searching for predicates in $KG$, we limit our scope to those connected to vertices in $\mathcal{R_V}(n^a, KG) \bigcup \mathcal{R_V}(n^b, KG)$. For each such vertex $v$, we retrieve all its connected predicates by using SPARQL queries \glabel{outgoingPredicate(v)} and \glabel{incomingPredicate(v)}. Note that we must check both directions because $PGP(q)$ is undirected.    

For each retrieved predicate $p$ and its corresponding description $d_p$, we use Equation~\ref{eq:similarity} to compute an affinity score $S(l_r, d_p)$, where $l_r$ is the label (see Definition~\ref{def:rPredicate} ) of $r$. Semantic match allows us to map, for instance, label ``wife'' to DPBedia predicate \glabel{dbo:spouse}.
We select the top-$k$ predicates, according to affinity score, as the set of \emph{relevant} predicates for $r$. Formally:

\begin{definition}[Relevant Predicates $\mathcal{R_P}$]
\label{def:rPredicate}

Given a question $q$ and a target knowledge graph $KG= (V, P)$ considered as a set of triples.
Let $r$ be a relation in $PGP(q)$ with label $l_r$ and let $n^a, n^b \in PGP(q)$ be the nodes connected to~$r$. 
Let $T_{rv} = \left\{v: \langle v, s_v\rangle \in \mathcal{R_V}(n^a, KG) \bigcup \mathcal{R_V}(n^b, KG) \right\}$ be the union of relevant vertices of $n^a$ and $n^b$.
The set of all outgoing and incoming predicates  connected to relevant vertices is $T_{pd} = \left\{ \langle p, d_p, v, o \rangle : v\in T_{vr},  
\RDFTYPE{$v$}{$p$}{?obj}\in KG \vee \RDFTYPE{?sub}{$p$}{$v$}\in KG \right\}$;~flag $o$ is $True$ if $v$ was an object in the triple, and labels $d_p$ are retrieved as explained above. 
Let $T_p = \left\{\langle p,S(l_r,d_p), v, o\rangle: \langle p, d_p, v, o \rangle \in T_{pd} \right\}$, where $S(\circ, \circ)$ is the semantic affinity score.
The set of relevant predicates $\mathcal{R_P}(r, KG)$  for relation $r$ in target $KG$, is the subset of $T_p$ that contains all pairs with the top-$k$ affinity scores. 
\end{definition}


\input{algorithms/relation_linking_algorithm}


For our running example $q^E$, the top-$5$ predicates for relation ``city on shore'', assuming the target KG is {\dbpedia}, are: \glabel{dbp:city}, \glabel{dbp:locationCity}, \glabel{dbo:nearestCity}, \glabel{dbp:cities} and \glabel{dbp:country}.

Algorithm~\ref{alg:relationLinking} summarizes the relation linking process for a single relation $r$; the algorithm must be called for every relation  $r \in PGP(q)$.
Line 2 constructs set $T_{rv}$ that contains the union of relevant vertices of $n^a$ and $n^b$, where $n^a$ and $n^b$ are the PGP nodes connected to $r$.
In lines 3-7, for every relevant vertex $v$ in $T_{rv}$, we execute at the target KG SPARQL queries \glabel{outgoingPredicate(v)} and \glabel{incomingPredicate(v)} to retrieve all predicates connected to $v$, together with their descriptions; we store the results into set $T_{pd}$. 
Then, lines 8-14 compute, for each predicate in $T_{pd}$, its semantic affinity to $r$, and store the resulting $\langle predicate, score \rangle$ pairs in $T_p$, refer to Definition~\ref{def:rPredicate}. In lines 10-12, we check if $?p$ is not human-readable, e.g., a sequence of random characters and digits, then we issue a query to fetch the description of $?p$. Finally, line 15 selects from $T_p$ the pairs with the top-$k$ score to construct the set of relevant predicates $\mathcal{R_P}(r, KG)$.    


\subsubsection{Complexity} 

Algorithm~\ref{alg:relationLinking} is executed $|\mathcal{R}|$ times, where $\mathcal{R}$ is the set of relations in $PGP(q)$; see Definition~\ref{def:pgp}. The most expensive part is the execution of the SPARQL queries. Each can be executed in constant time $c_{lk}$ as a look-up query in the indices~\cite{Hexastore, TripleBit} of the RDF engine. However, there are $2\cdot |T_{vr}|$ such queries, where $|T_{vr}|\leq 2\cdot maxVR$, since there are two nodes connected to $r$; refer to the definition of \glabel{potentialRelevantVertices} query. In practice, we use only $k < maxVR$ vertices, therefore the cost of lines 4-7 becomes $O(k\cdot c_{lk})$. The cost of the for-loop in line 9 depends on $T_{pd}$, which can potentially contain all predicates from the target KG. However, in practice we limit the number of retrieved predicates per vertex, therefore the cost is $O(k)$; the same applies to line 15.  
The resulting complexity of Algorithm~\ref{alg:relationLinking} is $O(k\cdot c_{lk}\cdot |\mathcal{R}|)$.

\begin{figure}[t]
  \centering
  \includegraphics[width=\columnwidth]{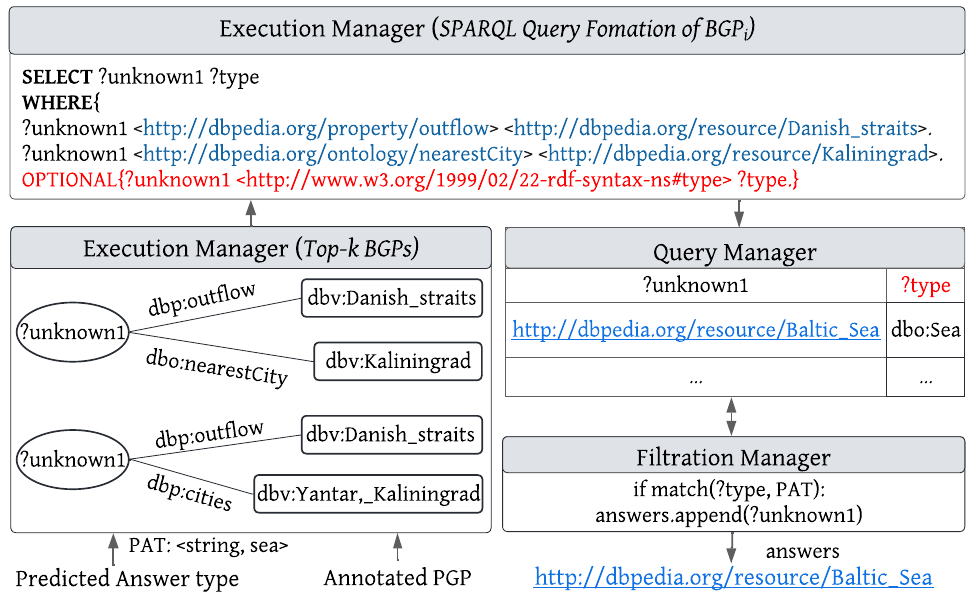}
\caption{The {\sysName} execution and Filtration phase uses the predicted answer type and the annotated PGP. {\sysName} generates a set of BGPs and ranks them to create up to $k$ queries. {\sysName} adjusts these queries with an optional triple pattern to return the type of the main unknown for filtering. 
}
\label{fig:exeandfilter}
\end{figure}

\subsection{Annotated Graph Pattern (AGP)}

Let $q$ be a question with a corresponding graph $PGP(q)$, constructed during question understanding. Given a knowledge graph $KG$, Algorithms~\ref{alg:entityLinking} and \ref{alg:relationLinking} generate, in a just-in-time manner, annotated graph $AGP(q,KG)$ that links the abstract components of $PGP(q)$ to actual vertices and predicates in $KG$, as shown in Figure~\ref{fig:architecture}. Formally: 

\begin{definition}[Annotated Graph Pattern (AGP)]
\label{def:pgp} 
Let $PGP(q)=(\mathcal{E}, \mathcal{R})$ be the phrase graph pattern for question $q$, and let $KG$ be the target knowledge graph. The corresponding annotated graph pattern $AGP(q, KG)$ is an undirected graph consisting of nodes $\mathcal{E}$ and relations $\mathcal{R}$, where every $n\in \mathcal{E}$ is annotated with relevant vertices $\mathcal{R_V}(n, KG)$, and every $r\in \mathcal{R}$ is annotated with relevant predicates $\mathcal{R_P}(r, KG)$. 
\end{definition}

\subsection{Semantic Affinity Calculation} 
\label{sec:semanticAffinity}
Here we discuss how to compute the semantic affinity score $S(\circ, \circ)$ that we use during entity and relation linking. Let $l=(l_1, l_2, ...)$ be a string consisting of a sequence of words. For each word $l_i \in l$, we generate a word embedding $E_w(l_i)$ using the FastText model~\cite{fasttext}, which is pre-trained on a large vocabulary of English words. The intuition is that, if two words are semantically similar, they will be close in the vector space of the embedding. If FastText cannot recognize $l_i$, then we generate a character embedding $E_c(l_i)$ instead, using the chars2vec~\cite{chars2vecGit, chars2vecArticle} pre-trained model, which captures the similarity of word spellings.    
Let $l^X$ be a string and let $X=(x_1, x_2, ..., x_{|l^X|})$ be an array of embeddings, where $x_i = E_w(l^X_i)$, if $l^X_i$ appears in FastText, or $x_i = E_c(l^X_i)$, otherwise. Let $l^Y$ also be a string, with its corresponding array of embedding $Y$. The semantic affinity \cite{similarity} between $l^X$ and $l^Y$, is defined as:   

\begin{equation}
    S(l^X, l^Y) = \frac{\sum_{x_i \in X, y_j \in Y} sim(x_i, y_j)}{|X| \cdot |Y|}
    \label{eq:similarity}
\end{equation}

\noindent where $sim(\circ, \circ)$ is the cosine similarity.
Equation~\ref{eq:similarity} considers all pairs of $(x_i, y_j)$. Some of these pairs may contain embeddings from different models, for instance $\left (E_w(l^X_i), E_c(l^Y_j)\right )$; in such cases, we define $sim(x_i, y_j)=0$.  

While the above is our default semantic affinity calculation method, we also experiment with sentence-based embedding models. We use the GPT-3~\cite{gpt} pre-trained transformer to generate a single embedding for the entire string. In this case, Equation~\ref{eq:similarity} is simplified as: $S(l^X, l^Y) = sim \left(E_{GPT}(l^X), E_{GPT}(l^Y)\right )$.

%% file: algorithms/entity_linking_algorithm.tex
\begin{algorithm}[t]
\caption{{\sysName}EntityLink}
\label{alg:entityLinking}
\begin{flushleft}

\textbf{Input:} $n$: a node in $PGP(q)$, $KG$: target knowledge graph, $maxVR$: max fetched vertices, $k$: number of vertices\\
\textbf{Output:} $n$ annotated with relevant k vertices $\mathcal{R_V}(n, KG)$\\
 
\end{flushleft}

\begin{algorithmic}[1]

    \If{$n.type$ is ``unknown''} \Comment{$n$ is a variable}
        \State \textbf{return} $n.\mathcal{R_V} \gets \varnothing $
    \EndIf
    \State $ T_d \gets $ potentialRelevantVertices($n.l_n, maxVR$) \Comment{SPARQL to  $KG$}
     \State $T_v \gets \varnothing$
     
     \For{every $\langle v, d_v\rangle \in  T_d$}
        \State $T_v \gets T_v \cup \langle v, S(n.l_n, d_v)\rangle$ \Comment{compute semantic affinity}
    \EndFor

    \State \textbf{return} $n.\mathcal{R_V} \gets$ all pairs from $T_v$ with top-$k$ affinity score 
\end{algorithmic}
\end{algorithm}

%% file: algorithms/relation_linking_algorithm.tex
\begin{algorithm}[t]
\vspace*{1ex}
\caption{{\sysName}RelationLink}
\label{alg:relationLinking}
\vspace*{-0.5ex}
\begin{flushleft}
\textbf{Input:} $r$: a relation in $PGP(q)$, $KG$: target knowledge graph, $k$: number of predicates \\
\textbf{Output:} $r$ annotated with relevant predicates $\mathcal{R_P}(r, KG)$\\ 
\end{flushleft}

\begin{algorithmic}[1]
    \State $n^a, n^b \gets$ $PGP(q)$ nodes connected to $r$
	\State $T_{rv} \gets$ set of all $v\in \mathcal{R_V}(n^a, KG) \cup \mathcal{R_V}(n^b, KG)$
    \State $T_{pd} \gets \varnothing$ \Comment{predicates of relevant vertices}
    \For{every $v\in T_{rv}$}
        \State $ T_{pd} \gets T_{pd}\cup$ outgoingPredicate($v$)      \Comment{SPARQL to  $KG$}
        \State $ T_{pd} \gets T_{pd}\cup$ incomingPredicate($v$)  
    \EndFor

    \State $T_p \gets \varnothing$

    \For{every $\langle p, d_p, v, o \rangle \in T_{pd}$}
        \If{not isHumanReadable($p$)} 
            \State $d_p \gets$ getPredicateDescription($p$) 
        \EndIf
        \State $T_p \gets T_p \cup \langle p, S(l_r, d_p), v, o\rangle$ \Comment{compute semantic affinity}
    \EndFor

\State \textbf{return} $r.\mathcal{R_P} \gets$ all pairs from $T_p$ with top-$k$ affinity score 
\end{algorithmic}
\vspace*{1ex}
\end{algorithm}

%% file: sections/execution_manager.tex
\section{{\sysName} Execution and Filtration}
\label{sec:exec}

Our JIT approach shifts the filtration process from RDF engines to {\sysName} and plans for it at runtime for a given question and an arbitrary KG. This shift helps {\sysName} to work without pre-processing.
Figure ~\ref{fig:exeandfilter} illustrates the execution and filtration in {\sysName}. The {\sysName} post-filtering method percolates answers collected by executing the semantically equivalent queries (BGPs) against $KG$, which is a set of RDF triples \RDFTYPE{sub}{$p$}{obj}. In $KG$, \glabel{obj} may be a certain class type for the vertex \glabel{sub} and $p$ will be an \glabel{rdf:type}\footnote{\url{https://www.w3.org/1999/02/22-rdf-syntax-ns\#type}} predicate. 
{\sysName} retrieves the class type of the main unknown, if available in $KG$, by extending the top selected BGPs by an optional triple pattern; \RDFTYPE{$unknown1$}{rdf:type}{$?c$}. 
Then, {\sysName} generates a SPARQL query from this set of adjusted BGPs, sends them for execution to the RDF engine, and maintains the set $\{\langle a, ?c\rangle\}$, where $a$ is a received answer and $?c$ is its associated class type. Finally, {\sysName} iterates over this set to filter out each $\langle a, ?c \rangle$, where $?c$ does not match \myNum{i} the predicted data type, in case of date, or numerical; or \myNum{ii} the predicted semantic type, if the predicted data type is string.\shorten

\input{algorithms/AnnotatedPGPToQuery}



\begin{definition}[Basic Graph Pattern (BGP)]
\label{def:BGP}
Let $AGP(q,KG)= (\mathcal{E}, \mathcal{R})$ be the annotated graph pattern for question $q$ and target knowledge graph $KG$. The set of triple patterns in $AGP(q,KG)$ is
$TP= \{\langle n^a, r, n^b\rangle : \langle n^a, r, n^b\rangle \in AGP(q,KG)\}$. 
For every triple in $TP$, assign a value $v^a \rightarrow n^a$, $v^b \rightarrow n^b$ and $p \rightarrow r$, such that
$v^a \in n^a.\mathcal{R_V},
v^b \in n^b.\mathcal{R_V}$ and
$p \in r.\mathcal{R_P}$.
The orientation $\langle v^a, p, v^b\rangle$, or $\langle v^b, p, v^a\rangle$ of the resulting triple, depends on flag $o$ of $p$ (see Definition~\ref{def:rPredicate}).


\end{definition}


Algorithm~\ref{alg:convertPgp} explains {\sysName}’s procedure for creating a list of ranked SPARQL queries from annotated graph pattern $AGP$. Line~1 creates $BGP_{all}$, a set of BGPs generated using all possible valid combinations of relevant vertices and predicates in $AGP$; note that Definition~\ref{def:BGP} describes only \emph{one} such BGP.  
In line 2, we calculate the score of each $BGP \in BGP_{all}$. The score of a $BGP$ is defined as:\shorten 

\begin{equation}
    score(BGP) = \frac{1}{|TP|} \sum_{\langle v^a, p, v^b\rangle \in TP} \left (s_{v^a} + s_{p} + s_{v^b} \right )
    \label{bgp_score}
\end{equation}

\noindent where $s_{v^a}$, $s_{v^b}$, and $s_{p}$ are the scores of the relevant vertices and predicates; refer to Definitions~\ref{def:rVertex} and \ref{def:rPredicate}. Line 3 sorts $BGP_S$ based on the calculated score. In line 4, we finally convert the top-$k$ BGPs in $BGP_R$ to their equivalent SPARQL queries. Each query is appended by an optional clause with triple pattern \RDFTYPE{$unknown1$}{rdf:type}{$?c$} to fetch the type or class linked to the main unknown, if any.\shorten

%% file: algorithms/AnnotatedPGPToQuery.tex
\begin{algorithm}[t]
\caption{Generate the Top-$k$ SPARQL queries from $AGP$}
\label{alg:convertPgp}
\begin{flushleft}
\textbf{Input:} $AGP(q,KG)$: an annotated graph pattern  \\
\textbf{Output:} $BGP_{SQ}$: Top-k SPARQL queries 
\end{flushleft}

\begin{algorithmic}[1]

    \State $BGP_{all} \gets  getBGPs(AGP(q,KG)) $ \Comment{all possible combinations}
   \State $BGP_S \gets calculateScores(BGP_{all})$ \Comment{using equation~\ref{bgp_score} }
  \State $BGP_R \gets rank(BGP_S)$
   \State $BGP_{SQ} \gets  getSPARQL(BGP_R,k)$ \Comment{with optional type clause}
\end{algorithmic}
\end{algorithm}

%% file: sections/results.tex
\section{Experimental Evaluation}
\label{sec:results}

\input{tables/preprocessing_time_stat}

\subsection{Evaluation Setup}

\subsubsection{Compared Systems}
We evaluate {\sysName} against gAnswer~\cite{gAnswer2014, gAnswer2018}, EDGQA~\cite{EDGQA}, and NSQA~\cite{NSQA}. 
EDGQA is state-of-the-art in both {\qald} and {\lcq}. gAnswer was ranked first in the QALD-9 challenge~\cite{qald9}. NSQA utilized deep learning models to support question understanding and linking and outperformed gAnswer in {\qald}. 
The code of both gAnswer~\cite{GAnswerGit} and EDGQA~\cite{EDGQAcode} is available. We reproduce the results of gAnswer and EDGQA. NSQA uses a logical neural network~\cite{LNN}, and special datasets for training the AMR model; both are not available. Thus, we only report the results provided at~\cite{NSQA}.\shorten

\subsubsection{Four Different Real KGs}
To evaluate these systems with diverse application domains, we use four real KGs, namely {\dbpedia}, YAGO~\cite{YAGOkgdownload, yago}, DBLP~\cite{DBLPrelease} and the Microsoft Academic Graph (\emph{MAG})~\cite{MAGrecords}. Both {\dbpedia} and YAGO are general-fact KGs, where most entities are about places and persons. DBLP and MAG are oriented to scientific publications, citations, authors and institutions, where some entities are identified by long phrases, such as a paper's title and a conference name. Moreover, these four KGs are of different sizes to study the effect of the graph size on pre-processing required by these systems, as illustrated in Table~\ref{tab:preprocessing}.

\subsubsection{Benchmarks and Question Sets}
QA benchmarks include English questions annotated with the corresponding SPARQL queries and the set of correct answers from a specific version of a KG. {\qald}~\cite{qald9} and {\lcq}~\cite{lcquad} are widely used to evaluate QA systems on different versions of {\dbpedia}. {\qald} has 408 questions as a training set, and 150 as a testing set. The average length of a question in {\qald} is 7.5 words. {\lcq} has a training set of 4000 questions and a testing set of 1000 questions. These questions are created based on different templates. 
There was no golden standard for YAGO, DBLP, or MAG. Thus, we asked a group of students with a background in computer science to express questions similar to {\qald}’s questions solved by most QA systems to find facts in YAGO, DBLP and MAG. We collected 100 questions per KG and annotated them with the correct SPARQL query and answers. These new benchmarks aim to evaluate the QA systems in solving similar questions to {\qald} in unseen KGs and domains. Table~\ref{tab:preprocessing} summarizes the five benchmarks we use. 
We used an evaluation metrics of: Precision (P), Recall (R), and Macro F1 (F1) and calculate them using the {\qald} automatic evaluation tool ~\cite{QALDscripts}.\shorten 

\subsubsection{RDF Engines for SPARQL Endpoints}
We use Virtuoso 7.2.5.2 as SPARQL endpoints, as it is widely adopted as an endpoint for big KGs, such as {\dbpedia} and MAG. We prepare five different Virtuoso endpoints for each of the benchmarks. The standard, unmodified installation of the Virtuoso engine was run at the endpoints and used by all QA systems in our experiments. The EDGQA linking method (Falcon) uses Elasticsearch. We use Elasticsearch 7.10.2 to index the KGs and enable the Falcon linking method to EDGQA.\shorten

\subsubsection{Computing Infrastructure}
We deploy each of the evaluated QA systems plus KGs on virtuoso SPARQL endpoints running on the same local machine. We use two different settings for our experiments. In the first setting, we use two Linux machines, each with 16 cores and 500GB RAM. These machines are used with all experiments except those related to MAG. In the second setting, we use \textit{five} Linux machines with \textit{32 cores} and \textit{3TB} RAM. EDGQA and gAnswer use these \textit{five} machines to pre-process MAG and generate the necessary indices for their approach to work.  

\subsubsection{{\sysName} implementation and settings}
{\sysName} is  implemented\footnote{\url{https://github.com/CoDS-GCS/KGQAn}} using \emph{Python} 3.7 and all models are trained using \emph{Pytorch} 1.11.0. Our implementation supports parallel execution and filtration via multi-threading. We did not enable multi-threading for the execution and filtration, as gAnswer and EDGQA do not have parallel support. We utilize BART~\cite{BART} for modelling our QU Seq2Seq model. Our semantic affinity model uses FastText's wiki-news-300d-1M model.
We also show experiments with GPT-3~\cite{gpt} for our QU and semantic affinity models in Sub-section~\ref{KGQAN-Analyzing}. 
{\sysName} needs four parameters: \myNum{i} \emph{Max Fetched Vertices} to decide how many vertices for a certain entity could be investigated,  \myNum{ii} \emph{Number of Vertices} to decide how many vertices could be used to annotate each node in the PGP, \myNum{iii} \emph{Number of Predicates} to decide how many predicates could be used to annotate each edge in the PGP, and \myNum{iv} \emph{Max number of Queries} to decide the number of equivalent SPARQL queries to be generated for the question. Our experiments use the following values \emph{400}, \emph{1}, \emph{20}, and \emph{40}, respectively. We decide the number of predicates based on the general average number of predicates per vertex. We tuned these parameters to work across different KGs, \textit{not to outperform for a specific one}.\shorten

\subsection{Experiments with Real KGs }
gAnswer, EDGQA, and NSQA are trained and evaluated using {\qald} and {\lcq}. We compare {\sysName} to these systems using five benchmarks on diverse real KGs. We performed the pre-processing required by gAnswer and EDGQA. Table~\ref{tab:preprocessing} summarizes the time and storage consumed by each system for pre-processing. We assessed these systems with seen, i.e., {\qald} and {\lcq}, and unseen benchmarks on YAGO, DBLP and MAG. Table~\ref{tab:qald_lcquad} shows the results of all systems on the five benchmarks. 

\subsubsection{Pre-processing Cost}
EDGQA uses an ensemble of three different linking methods, namely Falcon~\cite{falcon}, Dexter~\cite{dexter}, and EARL\cite{earl} for entity and relation linking. We use Falcon to index all KGs and enable EDGQA to work. To reproduce EDGQA’s results in {\qald} and {\lcq} we use the indices for Dexter and EARL, which are provided at the EDGQA repository. 
For MAG, we customize Falcon’s code with the right predicate describing each entity type, such as the paper's title for papers and author’s name for authors. gAnswer provided their indexing mechanism and relation mentions file that are needed in the pre-processing. We use the machines with \textit{3TB} of RAM to run the gAnswer indexing script. For EDGQA, Falcon consumes more time in indexing a KG than gAnswer's indexing method, as shown in Table~\ref{tab:preprocessing}.  

\subsubsection{Seen Benchmarks}
\input{tables/qald9_results_and_others}
For the {\lcq} benchmark, {\sysName} performs almost identically to EDGQA in F1 score and an improvement of 15\% in precision, as shown Table~\ref{tab:qald_lcquad}. EDGQA outperformed {\sysName} in the recall by almost 19\% due to the aggressive indexing of three different linking systems and the human-curated rules to understand questions of {\lcq}. These questions are automatically created using templates, i.e., curating rules per template helps perform well in training and testing questions. Unlike {\lcq}, {\qald} 's questions are created manually with different complexity. Thus, {\qald} is more challenging, e.g., curated rules may work with training questions and fail with testing questions. 
This explains the huge difference in EDGQA's F1 scores in {\lcq} and {\qald}. {\sysName} outperforms the existing QA systems in {\qald}, as shown Table~\ref{tab:qald_lcquad}, by achieving an impressive precision and F1 score of 49.81 and 43.99, respectively.

\subsubsection{Unseen Benchmarks on YAGO, DBLP and MAG} 
To evaluate the universality of the systems in processing a question against an arbitrary KG, we use three benchmarks unseen by all systems, including {\sysName}. One benchmark has questions similar to {\qald} and targets YAGO, a KG similar to {\dbpedia}. Another two benchmarks target a domain different from {\dbpedia}, where the questions find facts related to papers and authors in DBLP and MAG. Interestingly, gAnswer achieved better performance in the YAGO benchmark than the {\lcq} benchmark, as the questions in YAGO are similar to the ones in {\qald}. 
In MAG and most cases in DBLP, vertices' URIs end with code, e.g., ``Jim Gray'''s URI is mag$_e$:2279569217, see Subsection 5.1. The gAnswer inverted index is based on the URIs. So, gAnswer linking cannot find ``Jim Gray''. Thus, gAnswer answers two questions in DBLP and Zero in MAG due to its QU and linking. On DBLP and MAG, EDGQA fails to answer most questions, as it failed to understand questions about entities with longer phrases. 
{\sysName}’s question understanding model generalizes better in extracting entities with longer phrases. Moreover, {\sysName}’s semantic affinity model is trained based on general English text. This helps our JIT linking work well across KGs of different domains and utilize the RDF engines’ built-in indices.\shorten 

\begin{figure}
\vspace*{-2ex}
  \centering
  \includegraphics[width=\columnwidth]{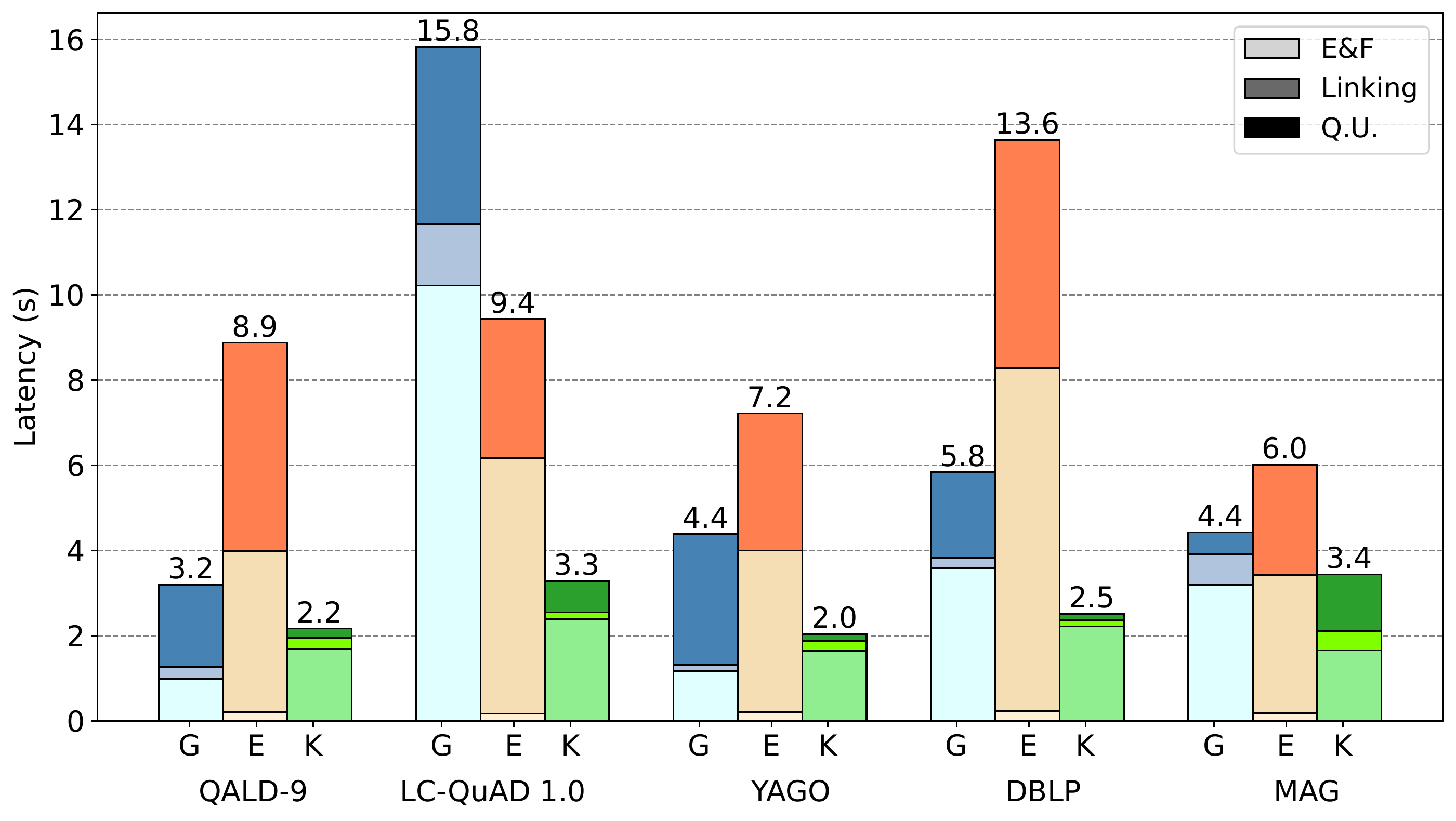}
\caption{Response time of gAnswer (G), EDGQA (E) and {\sysName} (K). Each bar shows average response time classified bottom-up into QU, Linking, and Execution/Filtration (E\&F).\shorten 
}
\label{fig:TimeQALDFigure}
\end{figure}

\subsubsection{Response time} 
This experiment analyzes the QA systems in terms of response time to a question. Per the system, we calculate the average response time for every set of questions. We use a local setting where the QA system and the benchmark SPARQL endpoint are deployed in the same machine to eliminate variability due to network latency or workload on the actual endpoint of KGs, such as MAG or {\dbpedia}. We report the average time of each step, question understanding ( QU), linking, and execution/filtration, as shown in Figure~\ref{fig:TimeQALDFigure}. The {\sysName} QU model consumes the majority of response time. The {\sysName} linking usually consumes the lowest time. Finally, the execution and filtration in {\sysName} consume more time than linking, depending on how complex the generated SPARQL queries are and the size of the query results.

In gAnswer and EDGQA, the total response time is dominated by the accuracy of question understanding in terms of: \myNum{i} the extracted entities and relations, which affect the linking time, and \myNum{ii} the number of extracted triples, which affects the number of generated SPARQL queries, i.e., execution time. 
The total response time is dominated by the complexity of the overall question-answering pipeline more than the graph size. For example, {\sysName} consumes similar time in {\lcq} and MAG. In general, EDGQA consumes more time in linking as the utilized linking methods use disk-based indices. In contrast, gAnswer loads its indices in memory before processing any question.\shorten

\begin{figure}
\vspace*{-2ex}
  \centering
  \includegraphics[width=\columnwidth]{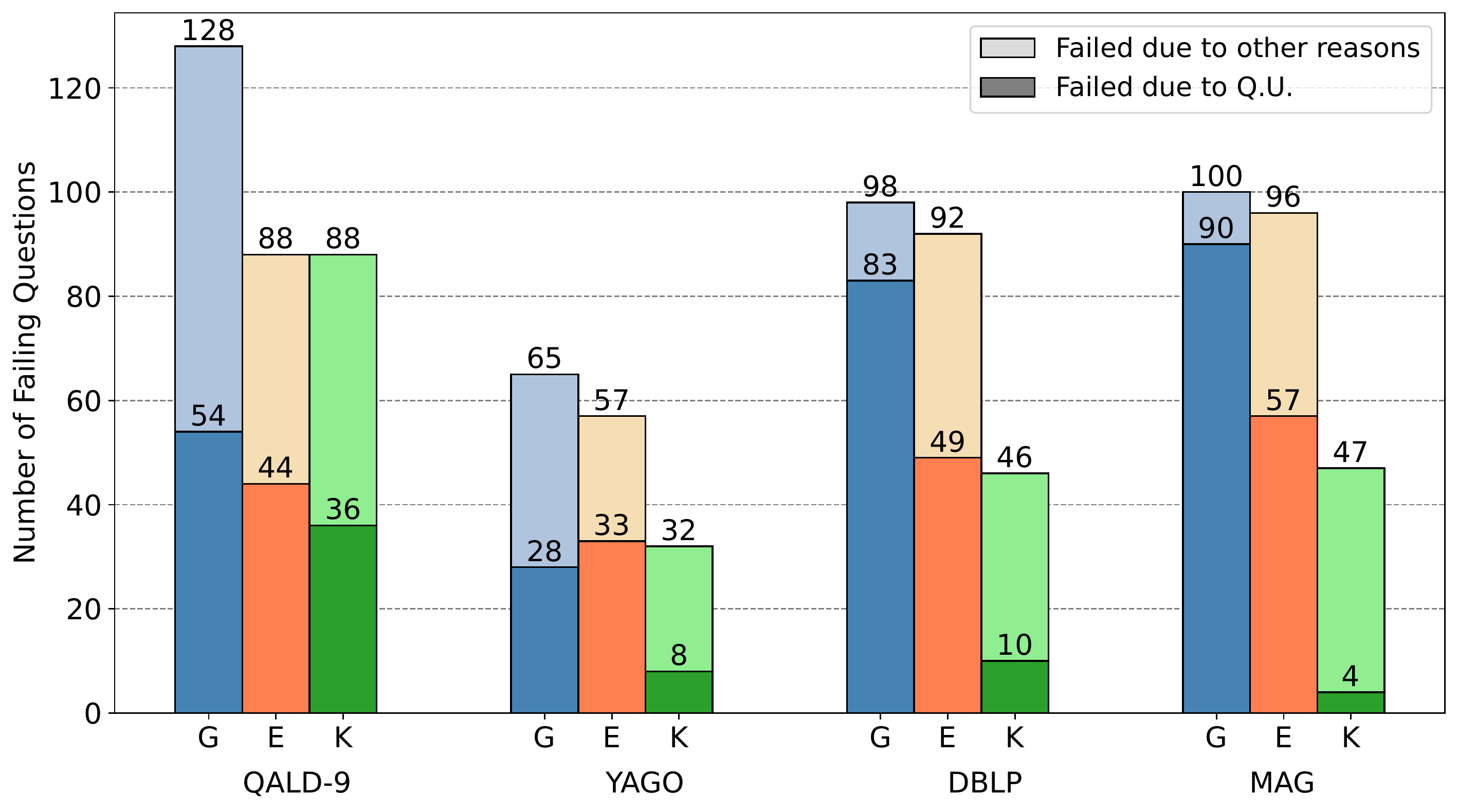}
\caption{Number of failing questions with R=0 and F1=0 in each benchmark.
Each bar shows the total number classified bottom-up into failing due to QU or others. {\sysName} fails in the least number of questions across diverse benchmarks.}
\label{fig:failedQuestion}
\end{figure}

\subsection{Analyzing {\sysName}}
\label{KGQAN-Analyzing}

\subsubsection{Question Understanding}
This experiment analyzes the QA systems in terms of total failure in answering a question, i.e., recall and F1 are zeros. This failure could be due to QU or other reasons, such as linking or filtering. Failing to understand a question means no entities or relation phrases are correctly detected. We manually count the failing cases due to QU in each benchmark, as summarized in Figure~\ref{fig:failedQuestion}. {\sysName} outperforms gAnswer and EDGQA in understanding most of the questions across benchmarks of different domains, as it fails in the least number of questions in total and due to QU per benchmark. 
{\sysName} was able to understand questions in an unseen domain, i.e., DBLP, better than gAnswer and EDGQA. 

\subsubsection{JIT Linking and Post-Filtering}
\begin{figure}[t]
\vspace*{-2ex}
  \centering
  \includegraphics[width=\columnwidth]{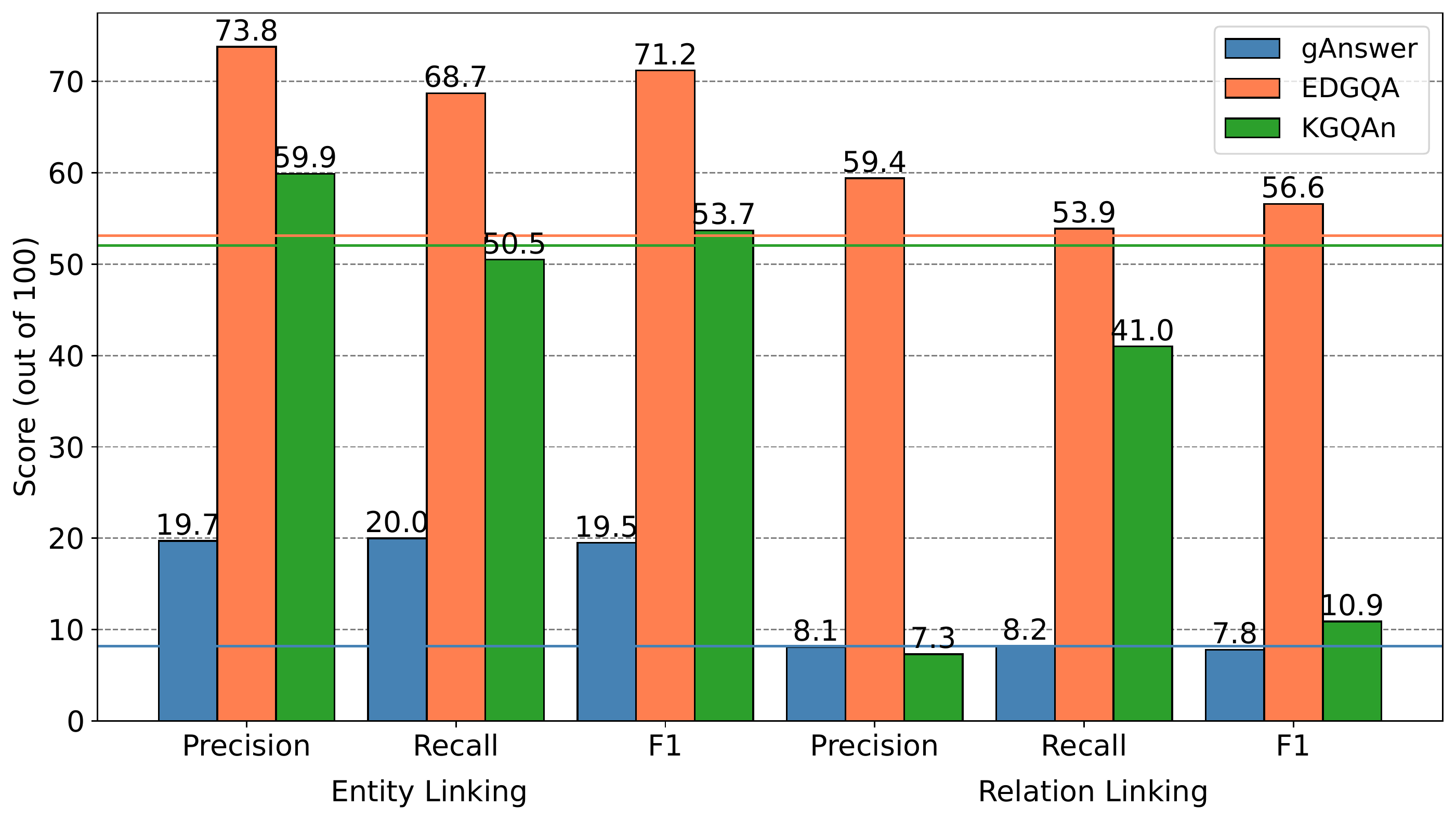}
  \vspace*{-3ex}
\caption{The entity and relation linking using {\lcq} show that EDGQA benefits from combining three linking methods. The horizontal lines show the final F1 score per system. 
Unlike EDGQA, {\sysName} achieves a final F1 score almost identical to its highest F1 in the entity linking.
}
\label{fig:LinkingComparison}
\end{figure}

A labeled dataset for the entity and relation linking task in {\lcq} is provided by \cite{earl}. In this experiment, we use this dataset to analyze the performance of the linking methods adopted by gAnswer and EDGQA, as shown in Figure~\ref{fig:LinkingComparison}. 
Our JIT linking approach eliminates the need for the pre-processing phase by offloading the linking task partially as queries executed by the RDF engine and followed by semantic ranking at {\sysName}. 
EDGQA utilizes three different linking systems. Thus, it achieves an outstanding performance in this task. gAnswer's QU is trained only on {\qald}. Therefore, it does not extract the correct entities and relations in most cases, and consequently, it does not perform well in the linking task in {\lcq}. Our approach aims at maximizing the recall to find the right vertex. 
{\sysName} performs a post-filtration of irrelevant answers to improve the precision. Hence, the final F1 score of {\sysName} is almost identical to the highest F1 score achieved in the entity linking. In contrast, the overall processing in EDGQA negatively affects the F1 score achieved by the three linking systems.\shorten

\subsubsection{Filtration effect}
\begin{figure}[t]
\vspace*{-2ex}
  \centering
  \includegraphics[width=\columnwidth]{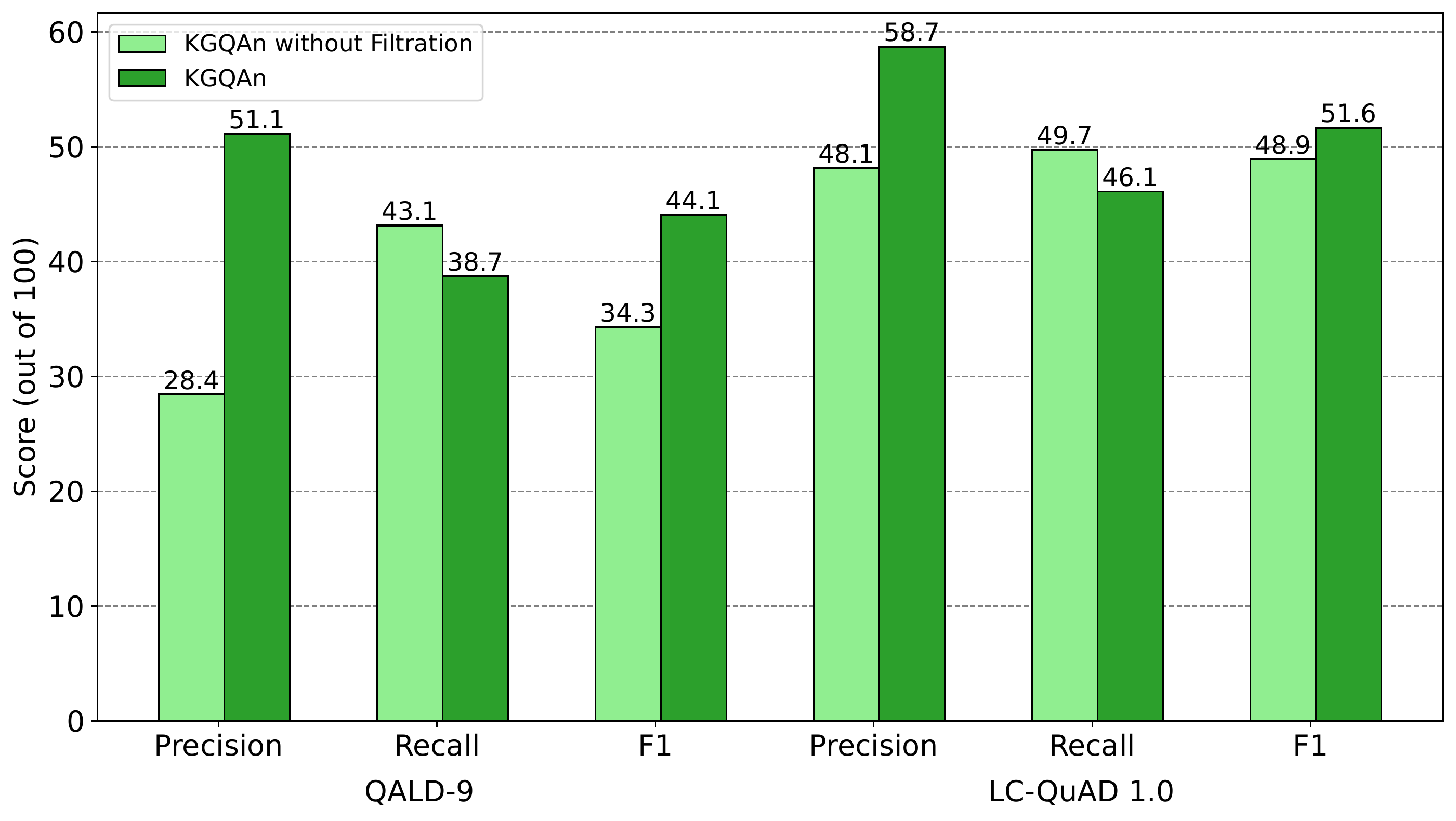}
  \vspace*{-2ex}
\caption{Analyzing {\sysName}'s performance with and without filtering, using {\qald} and {\lcq}. Filtering improves precision but slightly reduces recall. The resulting F1 score is improved for both benchmarks.}
\label{fig:FilteringComparison}
\end{figure}

This experiment analyzes {\sysName} performance with and without filtration.
For the lack of space, we show only our results on {\qald} and {\lcq}, as illustrated in Figure~\ref{fig:FilteringComparison}. 
{\sysName} predicts answer data type. 
Our model performs very well in filtering answers based on the data types date, numerical or boolean. Due to the high ambiguity, filtering answers using \textit{semantic} types is not as accurate as the data types date, numerical or boolean.  Overall, our filtration method is designed to avoid hurting the recall much. {\qald} has a higher ratio of questions whose answers are of type date, numerical or boolean than {\lcq}. Thus, {\sysName} with filtration achieved better in {\qald} than {\lcq}.

\input{tables/gpt3_ablation}

\subsubsection{{\sysName} and different pre-trained language models.}
This experiment analyzes the effectiveness of different pre-trained language models (PLMs) on the {\sysName} performance in both question understanding and semantic affinity,  which we use in both linking and filtration. For question understanding, we utilize BART~\cite{BART} and GPT-3~\cite{gpt} to model our triple patterns extraction task. 
For BART, we use the Huggingface API~\cite{huggingface} to perform the training, as it gives us the freedom to fine-tune the training parameters. Fine-tuning the GPT-3 model is only available through an OpenAI API~\cite{openai}. Thus, we had less control in optimizing our model using GPT-3. 
Our semantic affinity model depends on the representation (embeddings) of two sets of words. We develop a fine-grained approach to estimate the affinity at the granularity of a pair of words. For this approach, we use FastText and chars2vec models. As a coarse-grained approach, we use GPT-3 to get one vector embedding for each set of words.

We compare the performance of {\sysName} using these different combinations of models. 
{\sysName} by default utilizes our BART-based Seq2Seq and the fine-grained semantic affinity (FG) models. We create two different variations of {\sysName} where we replace our BART-based Seq2Seq model in one variation with our model trained using GPT-3 for question understanding. In the second variation, we use our BART-based Seq2Seq model and the coarse-grained semantic affinity (CG) based on GPT-3. Table~\ref{tab:languagemodels} illustrates the final F1 score for each benchmark using {\sysName} based on the default setting, and the two variations, from left to right. Our default setting outperforms the two variations in most cases.

\subsection{Taxonomy of benchmark questions}
\input{tables/taxonomyTables/taxonomy}
We develop a taxonomy to study the complexity of the benchmark questions based on different characteristics, and evaluate the performance of {\sysName}, gAnswer and EDGQA based on this taxonomy. 
As illustrated in Table~\ref{tab:maintaxonomy}, our taxonomy categorizes the complexity of the questions based on \myNum{i} the shape of the candidate SPARQL query and \myNum{ii} the linguistic complexity. A SPARQL query could be classified into a star or path query~\cite{queryshapes}. In our taxonomy, a star query is a query consisting of one or more triple patterns sharing the same subject. A path query contains at least two triple patterns where an object in one triple pattern is a subject in another pattern, i.e., a linear shape. Based on the linguistic classification provided by LC-QuAD 2.0~\cite{lcquad2}, the questions in {\qald} and the unseen benchmarks are classified into single fact, single fact with type, multi-facts, and booleans. {\sysName} outperforms gAnswer and EDGQA in most cases across the different benchmarks.

%% file: tables/preprocessing_time_stat.tex
\begin{table*}[t]
 \caption{The used benchmarks, size of each KG, and time taken by the QA systems for pre-processing, i.e., indexing the KG.}
 \label{tab:preprocessing}
        \begin{tabular}{lccc|cc|cc}
        \hline
        \multicolumn{4}{c}{\textbf{Benchmark Statistics}}&\multicolumn{2}{|c}{\textbf{EDGQA - Indexing by Falcon}} & \multicolumn{2}{|c}{\textbf{gAnswer}} \\
        \textbf{Benchmarks}&\textbf{\#Questions} & \textbf{KG Name} & {\textbf{\#Triples (M)}} & {\textbf{Index Time (hrs)}} &{\textbf{Index Size (G)}} &{\textbf{Index Time (hrs)}} &{\textbf{Index Size (G)}}\\
        \hline
        {\qald} &  150 & {\dbpedia}-10 & 194 & 6.51 & 1.80 & 2.86 & 8.60\\
        {\lcq} & 1000 & {\dbpedia}-04 & 140 & 6.23 & 1.70 & 2.28 & 6.60\\ 
        YAGO-Bench & 100  & YAGO-4 & 145 & 6.88 & 2.00 & 1.81 & 4.10\\ 
        DBLP-Bench & 100 & DBLP& 136 & 4.83 & 1.60 & 1.91 & 5.20 \\ 
        MAG-Bench & 100 & MAG& 13000 & 103.22 & 92.00 & 37.40 & 319.00\\ \hline
\end{tabular}
\end{table*}

%% file: tables/qald9_results_and_others.tex
\begin{table*}[t]
 \caption{Results of five benchmarks. {\sysName}'s recall and precision across KGs of different domains proves that {\sysName} performs better than other systems due to our Seq2Seq model and JIT linking that work without a pre-processing phase for a specific KG. 
 }
 \label{tab:qald_lcquad}
        \begin{tabular}{lcccccc|ccccccccc}
        \hline
         & \multicolumn{3}{c}{\textbf{{\qald}-{\dbpedia}}} & \multicolumn{3}{c}{\textbf{{\lcq}-{\dbpedia}}} & 
         \multicolumn{3}{|c}{\textbf{{YAGO}}} & \multicolumn{3}{c}{\textbf{{DBLP}}} & \multicolumn{3}{c}{\textbf{{MAG}}}  \\
        \textbf{System}&{\textbf{P}} &{\textbf{R}} &{\textbf{F1}} &{\textbf{P}} &{\textbf{R}} &{\textbf{F1}} 
        &{\textbf{P}} &{\textbf{R}} &{\textbf{F1}} &{\textbf{P}} &{\textbf{R}} &{\textbf{F1}} &{\textbf{P}} &{\textbf{R}} &{\textbf{F1}} \\
        \hline
        \color{darkgray}{NSQA}
        & \color{gray}{31.89} &  \color{gray}{32.05}  &  \color{gray}{31.26}  &  \color{gray}{44.76}              &  \color{gray}{45.82} &  \color{gray}{44.45} &  
        -    &    -    &    -  &    -    &    -    &    -  &    -    &    -    &    -\\ 
        gAnswer & 29.34 & 32.68  & 29.81 &  82.21  & 4.31  &    8.18  &    
        \textbf{58.49}   &   34.05   &   43.04  & \textbf{78.00} & 2.00 &   3.90 &    0.0    &    0.0    &    0.0 \\
        EDGQA   & 31.30  & \textbf{40.30} & 32.00 & 50.50 & \textbf{56.00}  & \textbf{53.10} &    
       41.90 &   40.80   &    41.40  &    8.00    &   8.00   &    8.00  &   4.00   &    4.00    &    4.00 \\  \hline
        \textbf{{\sysName}}    & \textbf{51.13}  & 38.72 &  \textbf{44.07}   & \textbf{58.71}   &  46.11    &  51.65 &   
        48.48  & \textbf{65.22}&  \textbf{55.62} &  57.87    &   \textbf{52.02}   & \textbf{54.79} &   \textbf{55.43}   & \textbf{45.61}  &    \textbf{50.05}\\ \hline 
        \end{tabular}
\end{table*}

%% file: tables/gpt3_ablation.tex
\begin{table}[t]
 \caption{{\sysName}'s performance using different pre-trained models for training our QU and semantic affinity (SA) models. Our default settings using BART and fine-grained (FG) affinity achieve better F1 scores in most cases than the variations with GPT-3 in both QU and the coarse-grained (CG) affinity.\shorten
 }
  \label{tab:languagemodels}
  \centering
\begin{tabular}{lccc}
\hline
 \multicolumn{1}{c}{\textbf{}}& 
 \multicolumn{1}{c}{\textbf{QU: BART}}&
 \multicolumn{1}{c}{\textbf{QU: GPT-3}}&
 \multicolumn{1}{c}{\textbf{QU: BART}}\\
  \multicolumn{1}{c}{\textbf{Benchmarks}}& 
 \multicolumn{1}{c}{\textbf{SA: FG}}&
 \multicolumn{1}{c}{\textbf{SA: FG}}&
 \multicolumn{1}{c}{\textbf{SA: GPT-3}}\\
        \hline
        {\qald} & \textbf{44.07} & 42.12 & 42.60 \\
        {\lcq} & 51.65 & \textbf{52.87} & 50.86\\
        YAGO & \textbf{55.62} & 54.94 & 55.02 \\
        DBLP & \textbf{54.79}  & 54.42 & 41.72\\ 
        MAG &  \textbf{50.05} & 49.26 & 37.64\\ 
        \hline
\end{tabular}
\end{table}

%% file: tables/taxonomyTables/taxonomy.tex
\begin{table*}[t]
 \caption{Number of questions solved by the three systems compared using the taxonomy of the benchmark questions which depends on the SPARQL query shapes and the linguistic complexity of the questions according to LC-QuAD 2.0~\cite{lcquad2}}
\label{tab:maintaxonomy}
 \centering
\begin{tabular}{l|rrrr|rrrr|rrrr|rrrr|rrrr|rrrrl}

\cline{1-25}

 \multicolumn{1}{c}{\textbf{Query type}}&
 \multicolumn{8}{|c}{\textbf{SPARQL shape}}&\multicolumn{16}{|c}{\textbf{LC-QuAD 2.0 taxonomy}}\\
 \cline{2-25}

 &
 
 \multicolumn{4}{c|}{\textbf{Star}}                                                                                                   & \multicolumn{4}{c|}{\textbf{Path}}                                                                                                                                    & \multicolumn{4}{c|}{\textbf{Singe fact}}                                                                                                                              & \multicolumn{4}{c|}{\textbf{Fact with type}}                                                                                                                          & \multicolumn{4}{c|}{\textbf{Multi fact}}                                                                                                                              & \multicolumn{4}{c}{\textbf{Boolean}}                                                                                                                                  &   \\
\textbf{Benchmark}            & \begin{sideways}\# queries\end{sideways} & \begin{sideways}\textbf{KGQAn}\end{sideways} & \begin{sideways}EDGQA\end{sideways} & \begin{sideways}gAnswer\end{sideways} & \begin{sideways}\# queries\end{sideways} & \begin{sideways}\textbf{KGQAn}\end{sideways} & \begin{sideways}EDGQA\end{sideways} & \begin{sideways}gAnswer\end{sideways} & \begin{sideways}\# queries\end{sideways} & \begin{sideways}\textbf{KGQAn}\end{sideways} & \begin{sideways}EDGQA\end{sideways} & \begin{sideways}gAnswer\end{sideways} & \begin{sideways}\# queries\end{sideways} & \begin{sideways}\textbf{KGQAn}\end{sideways} & \begin{sideways}EDGQA\end{sideways} & \begin{sideways}gAnswer\end{sideways} & \begin{sideways}\# queries\end{sideways} & \begin{sideways}\textbf{KGQAn}\end{sideways} & \begin{sideways}EDGQA\end{sideways} & \begin{sideways}gAnswer\end{sideways} & \begin{sideways}\# queries\end{sideways} & \begin{sideways}\textbf{KGQAn}\end{sideways} & \begin{sideways}EDGQA\end{sideways} & \begin{sideways}gAnswer\end{sideways} &   \\ 
\cline{1-25}
QALD-9               & 131                                      & {\textbf{60}}                                  &   56                                  & 21                                      & 19                                       & 2                                   &      \textbf{5}                               & 0                                      & 81                                       & \textbf{46}                                 & 41                                    & 16                                      & 28                                       & 7                                   &  \textbf{8}                                   &   3                                    & 37                                       & 9                                   &     9                                &               2                        & 4                                        & 0                                   &        \textbf{3}                             &                 0                      &   \\
YAGO-B               & 92                                       & \textbf{63}                                  & 39                                  & 32                                    & 8                                        & \textbf{5}                                   & 4                                   & 3                                     & 87                                       &\textbf{61}                                 & 40                                  & 33                                    & 6                                        & \textbf{5}                                   &  3                                   &     2                                  & 6                                        & \textbf{2}                                   &      0                               &  0                                     & 1                                        & 0                                   &   0                                  &                                       0&   \\
DBLP-B               & 92                                       & \textbf{46}                                & 8                                     &           2                            & 8                                       & \textbf{8}                                   &             0                        &                        0               & 85                                       & \textbf{49}                                  &     8                                &  1                                     & 11                                       & \textbf{4}                                   &                            0         &                                    1  & 4                                        & \textbf{1}                                   & 0                                    &   0                                    & 0                                  & -                                   &     -                                &             -                          &   \\
MAG-B                & 77                                       & \textbf{44}                                  & 4                                   & 0                                     & 23                                       & \textbf{9}                                   & 0                                   & 0                                     & 75                                       & \textbf{40}                                  &              4                       &                         0              & 7                                        & \textbf{2}                                   &             0                        &                        0               & 16                                       & \textbf{9}                                   & 0                                    &            0                           & 2                                        & \textbf{2}                                   & 0                                    &   0                                    &   \\ 
\cline{1-25}

\end{tabular}
\end{table*}

        


%% file: sections/related_work.tex
\section{Related Work }
\label{sec:relatedwork}

There is a growing effort to develop QA systems based on different techniques for question understanding and linking~\cite{Fatma2020, Fatma2O22}. QA systems are classified into SPARQL- and non-SPARQL-based. Examples of non-SPARQL-based systems are QAmp~\cite{QAMP}, Treo~\cite{Treo}, and others~\cite{YaoD14, WuFofe}. These systems do not translate questions into SPARQL queries. Instead, they implement proprietary query engines; therefore, they cannot be used with arbitrary SPARQL endpoints. 

SPARQL-based systems translate natural language questions into SPARQL queries using various techniques for: \myNum{i} question understanding, to generate an intermediate abstract representation; and \myNum{ii} linking the abstract representation to vertices and predicates of a KG. Examples include gAnswer~\cite{gAnswer2014, gAnswer2018}, EDGQA~\cite{EDGQA}, NSQA~\cite{NSQA}, WDAqua-core1~\cite{WDAquaWWW18}, Bio-SODA \cite{biosoda}, and QAnswerKG~ \cite{qanswerKG}. 

The  question understanding techniques are classified into: 
\myNum{i} rule-based, such as gAnswer~\cite{gAnswer2014, gAnswer2018}, EDGQA~\cite{EDGQA}, and  WDAqua-core1~\cite{WDAquaWWW18}. They curate rules based on Part-of-Speech (POS) tagging to extract entities and relation phrases. Unlike {\sysName}, systems based on human-curated rules are hard to generalize to a broad set of applications and groups of diverse users.  
\myNum{ii} deep learning-based, such as NSQA~\cite{NSQA}, which curates tree representations of a large set of questions and uses deep learning to train a semantic parser called AMR. Note that, AMR has different granularity than SPARQL; thus, there is a need to adjust NSQA to work with new domains~\cite{NSQA}. {\sysName} does not suffer this  issue, as SPARQL queries are generated by traversing the PGP, whose nodes and edges are annotated by relevant vertices and predicates in a KG.  
\myNum{iii} token-based techniques~\cite{biosoda, qanswerKG} that extract the longest sequence of keywords in a question that matches an entry in an inverted index built by these systems. Token-based systems are  KG specific.

{\sysName} annotates the generated PGP using our just-in-time approach that does not demand any pre-processing of the target KG. Unlike {\sysName}, existing systems index the target KG in advance, such as EDGQA~\cite{EDGQA} and gAnswer~\cite{gAnswer2014, gAnswer2018}, or build a vector space of all KG's vertices and predicates to predict the linking based on deep learning models, such as NSQA~\cite{NSQA}. In these systems, the pre-processing cost is proportional to the KG's size. For example, in gAnswer, the pre-processing complexity is polynomial to the number of KG's vertices~\cite{gAnswer2018, gAnswer2014}. EDGQA utilizes three different systems to index the target KG, where each system indexes the KG differently. 

{\sysName} extracts a sequence of triple patterns from a question; entities can be  unknowns (i.e., variables). 
Unlike our problem, existing relation triple extraction techniques~\cite{triples2020, triples2022} do not deal with unknowns; furthermore, relations are chosen from a pre-defined list, that is, they are not extracted from the given text. Thus,  relation triple extraction models and their training datasets cannot be used for our task. 

We train our task as a Seq2Seq model using BART~\cite{BART}. 
Existing systems also use Seq2Seq models.
For example, Ref.~\cite{IBMIWSC2021} is trained with extracted KG-specific information to generate a sequence of predicates for relations in a question.
Also, Ref.~\cite{WangTWK20} gets an annotated question and generates an annotated SQL query.
QAmp~\cite{QAMP} also utilizes a Seq2Seq model, not as a genration task, but to label sequences in a question as an entity, predicate, or class.
Unlike these models, {\sysName}'s Seq2Seq model does not depend on a query language, a specific KG, or a particular domain. Thus, our model has more flexibility to understand questions in different domains without needing a domain expert to prepare a training set.

%% file: sections/conclusion.tex
\section{Conclusion}
\label{sec:conc}
Existing QA systems are domain- and KG-specific and demand an expensive pre-processing phase. Thus, they cannot be used on-demand to process a question against an arbitrary KG. This paper presents {\sysName}, an on-demand KG question-answering platform that overcomes these limitations. {\sysName} proposes a novel formalization of question understanding as a triple pattern extraction modelled using a Seq2Seq neural network. Our model generalizes to understand questions across diverse domains. Moreover, {\sysName} introduces a just-in-time linking and filtering approach, which performs entity and relation linking as semantic search queries partially offloaded to the RDF engines. 
We  evaluate the state-of-the-art (SOTA) QA systems using broadly utilized benchmarks and four diverse real-life KGs. Based on the KG size, {\sysName} saves a few hours to days of pre-processing. {\sysName} achieves comparable F1 score to the SOTA for the common benchmarks, but outperforms the SOTA by a large margin for previously unseen KGs.
In our future work, we plan to expand the our question understanding training to include more complex questions; we also plan to support multi-intention questions.

\nsstitle{Acknowledgement.}
We thank the authors of EDGQA, gAnswer, and Falcon for their assistance to reproduce their results. 